\documentclass[acmtog,nonacm]{acmart}
\renewcommand\footnotetextcopyrightpermission[1]{}
\renewcommand\authorsaddresses{} 
\AtBeginDocument{%
  }

\usepackage[T1]{fontenc}    
\usepackage{hyperref}       
\usepackage{url}            
\usepackage{booktabs}       
\usepackage{algorithm}
\usepackage{algpseudocode}
\usepackage{nicefrac}       
\usepackage{microtype}      
\usepackage{xcolor}         
\usepackage{graphicx}
\usepackage{listings}       
\usepackage{float}
\newfloat{listing}{tbp}{lop}
\floatname{listing}{Listing}
\usepackage{caption}

\usepackage[table]{xcolor}
\usepackage{array}          
\usepackage{longtable}      
\usepackage{multirow}
\usepackage{rotating}       
\usepackage{caption}        

\usepackage{xspace}
\newcommand{\flip}{\protect\reflectbox{F}LIP\xspace }

\begin{document}

\title{MRD: Using Physically Based Differentiable Rendering to Probe Vision Models for 3D Scene Understanding}

\author{Benjamin Beilharz}
\affiliation{
	\institution{Perception Lab, Centre for Cognitive Science, Technical University of Darmstadt}\\
	\institution{Center for Mind, Brain and Behavior (CMBB),  Universities of Marburg, Giessen, and Darmstadt}
	\country{Germany}
}
\email{benjamin.beilharz@tu-darmstadt.de}
\author{Thomas S. A. Wallis}
\affiliation{
	\institution{Perception Lab, Centre for Cognitive Science, Technical University of Darmstadt}\\
	\institution{Center for Mind, Brain and Behavior (CMBB),  Universities of Marburg, Giessen, and Darmstadt}
	\country{Germany}
}
\email{thomas.wallis@tu-darmstadt.de}
\renewcommand{\shortauthors}{Beilharz et al.}

\begin{abstract}
	While deep learning methods have achieved impressive success in many vision benchmarks, it remains difficult to understand and explain the representations and decisions of these models. Though vision models are typically trained on 2D inputs, they are often assumed to develop an implicit representation of the underlying 3D scene (for example, showing tolerance to partial occlusion, or the ability to reason about relative depth). Here, we introduce \textbf{MRD} (metamers rendered differentiably), an approach that uses physically based differentiable rendering to probe vision models’ implicit understanding of generative 3D scene properties, by searching for 3D scene parameters that are physically different but produce the same model activation (i.e. are model metamers). Unlike previous pixel-based methods for evaluating model representations, these reconstructions are always grounded in physical scene descriptions, so we can, for example, probe a model's sensitivity to object shape while holding material and lighting constant.
	Making physically based differentiable rendering converge against a deep-network objective is itself a substantial technical challenge, and we present MRD as a proof of concept.
	We separate two outcomes: whether MRD finds a model metamer (our primary criterion, defined relative to a calibrated baseline reconstruction), and to what extent the optimization recovers the ground-truth scene parameters (a secondary outcome).
	Across multiple networks, MRD reliably finds material metamers, but these are frequently not the ground-truth material. This indicates that the networks' equivalence classes for material are rather broad. For shape, reconstructions often fail to meet any of our metamer criteria, although clear success cases exist (e.g. in particular for simple geometries).
	These reconstructions can be used to investigate the physical scene attributes to which models are sensitive or invariant.
	MRD holds promise for advancing our understanding of both computer and human vision, enabling us to ask how physical scene parameters cause changes in model responses.
\end{abstract}
\begin{teaserfigure}
	\includegraphics[width=\textwidth]{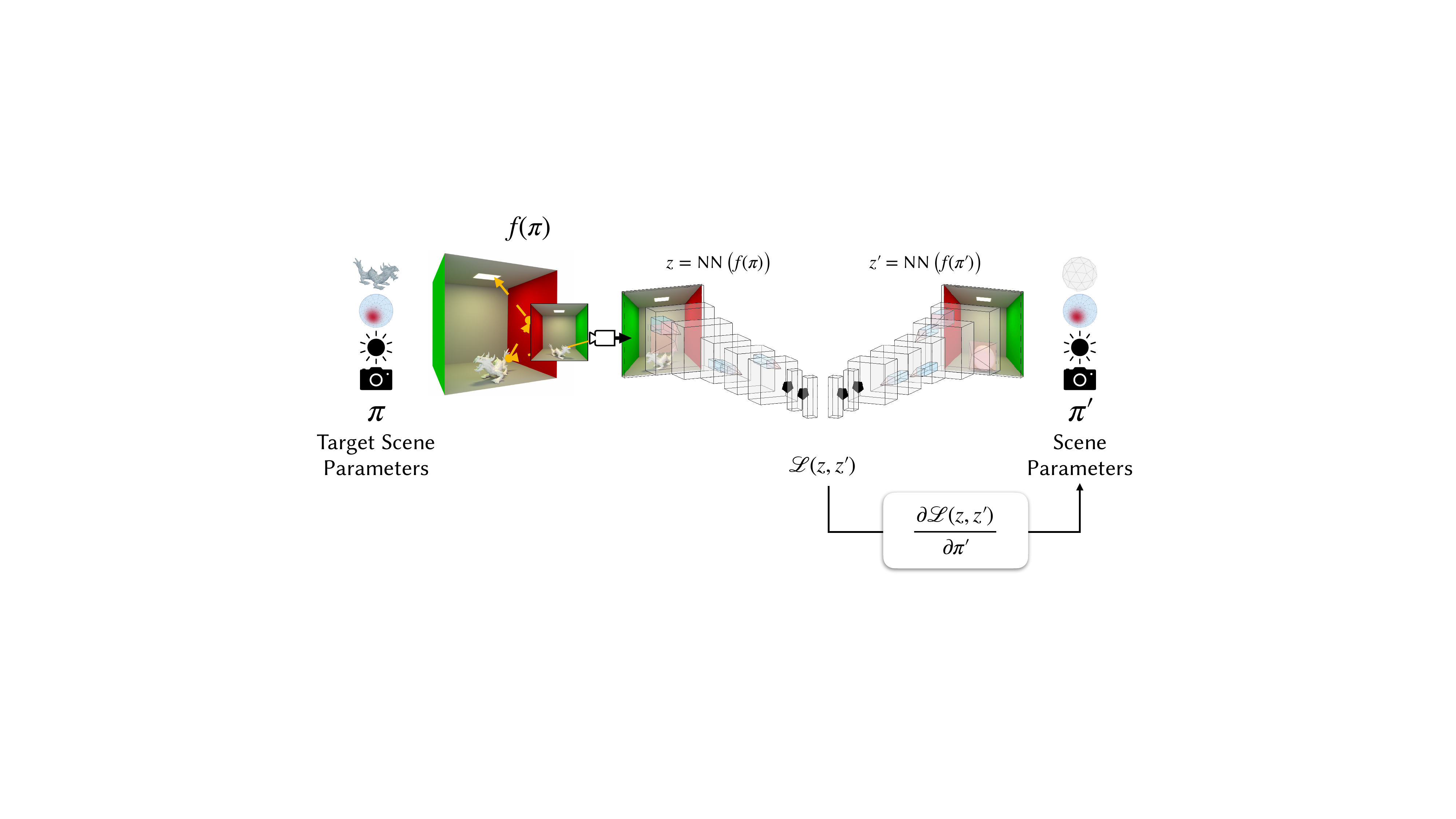}
	\caption{An overview of metamers rendered differentiably (MRD). Starting from a scene with known parameters $\pi$, we sample camera positions on the unit sphere and render a set of views using our differentiable rendering function \( f(\pi)  \). A new scene is initialized from another state $\pi^\prime$ (for example, a sphere shape instead of a dragon), rendered by \( f(\pi^\prime) \). The key step is that both image sets --- from ground-truth and initialized parameters --- are passed through a given frozen neural network \( \mathrm{NN}(\cdot) \) to obtain latent representations \( z \) and \( z^\prime \); the optimization objective is the distance (e.g. mean squared error) between these latents, not between pixels. We compute the gradient of this latent loss with respect to the scene parameters and backpropagate through the renderer, updating the target scene parameters (here, geometry) while holding other parameters (e.g. lighting) constant. This enables targeted probing of a neural network's understanding of scene properties by separating physical causes, and can uncover invariance or even equivalence classes.}
\end{teaserfigure}
\maketitle
\section{Introduction}\label{sec:introduction}

Deep learning has revolutionized pattern recognition from visual input.
Image-computable models can now perform many tasks with performance matching or exceeding humans, and their activations can correlate highly with visually-driven responses in primate brains \cite{doerigNeuroconnectionistResearchProgramme2023}.
However, it remains difficult to explain how and why these models make the decisions they do. Some work \cite{aliciogluSurveyVisualAnalytics2022, zimmermann2021how} probes whether models truly understand scenes, and judge whether these explanations might also provide explanations of visual processing in humans and other animals \cite{neriDeepNetworksMay2022,funkeFivePointsCheck2021,wichmannAreDeepNeural2023}.

In this work, we demonstrate how a relatively new technology from computer graphics --- physically based differentiable rendering (PBDR) --- can be used to evaluate 3D scene understanding in image-computable vision models.
PBDR allows the reconstruction of physically plausible 3D scene parameters, such as geometry, camera parameters, and material definitions, via optimization with gradient descent.
In contrast to approaches using neural networks for inverse rendering \cite{mildenhallNeRFRepresentingScenes2020,lingNeRFNonDistantEnvironment2024,lopesMaterialTransformsDisentangled2024,lingNeRFNonDistantEnvironment2024,liangGSIR3DGaussian2024}, the PBDR approach is always grounded in the physics of light transport, allowing the physical causes of the image to be separated, decomposed and thereby understood.

Applying PBDR allows one to synthesize new physical scene descriptions which, for example, cause matching model activations to a target scene but are physically different (i.e. are \textit{model metamers}; \cite{featherModelMetamersReveal2023,featherMetamersNeuralNetworks2019}).
This objective has long been used in the study of human vision, first to support trichromatic color theory \cite{maxwellXVIIIExperimentsColourPerceived1857}, and then for more general image representations \cite{freemanMetamersVentralStream2011,wallisTestingModelsPeripheral2016, balasSummarystatisticRepresentationPeripheral2009,wallisImageContentMore2019},  because it allows the identification of perceptual invariants.
Here, we combine PBDR with the metamerism objective to create metamers rendered differentiably (\textbf{MRD}).
Rather than trying to infer what the model might understand by interpreting noisy pixel images (e.g. in existing synthesis-based explanation methods), the user can interpret vision neural networks by using model representations to reconstruct specific scene parameters, represented in the physical units of the generating scene.
This also opens possibilities to fine-tune existing models on specific scene properties.

To demonstrate the usefulness of PBDR as a tool for model interpretability, we evaluate the implicitly-learned 3D knowledge of vision models trained on 2D images.
Vision models are assumed to learn about the underlying 3D structure of scenes, even by training only on 2D images.
For example, models trained on 2D scenes can perform well on tasks such as novel view synthesis \cite{mildenhallNeRFRepresentingScenes2020}, depth estimation \cite{godard2017unsupervised}, and 3D object reconstruction \cite{groueix2018papier, wang2018pixel2mesh}.
We demonstrate MRD in two example settings: 
first, we investigate the general problem of ``material appearance" by investigating the recovery of surface properties (bidirectional scattering distribution function, BSDF).
Second, we examine the recovery of shape (geometry) in the \textit{Learned Perceptual Image Patch Similarity (LPIPS)} \cite{zhangUnreasonableEffectivenessDeep2018} metric and in \textit{ImageNet} trained networks such as \textit{ResNet-50} \cite{heDeepResidualLearning2015} and its shape-bias induced version \textit{ResNet-50 SIN} trained on a stylized version of \textit{ImageNet} \cite{geirhosIMAGENETTRAINEDCNNSARE2019}.
We also present results for 
\textit{CLIP} \cite{radfordLearningTransferableVisual2021} as a common multi-model embedding backbone.

Our specific contributions are (1) a new method to understand learned visual representations of neural networks by linking their activations to physical environmental properties and efficiently optimizing to find invariants, and (2) a proof-of-concept evaluation of contemporary vision models using this method. We stress the proof-of-concept framing: making physically based differentiable rendering converge against a deep-network objective is a substantial technical challenge, and our results map out where it succeeds and where it does not. Concretely, MRD reliably finds material metamers (which are often not the ground-truth material, revealing broad equivalence classes), whereas shape reconstructions often fail to satisfy our metamer criteria, with informative exceptions.
Because it allows the decomposition of model activations into physical causes, we hope that this method will become an important tool for evaluating learned visual representations.

\subsection{Related work}\label{sec:related_work} 
\textbf{Synthesis-based explanation methods} have a long history in vision science and computer vision.
The central idea is to understand the representations of image-computable vision models by synthesizing new images via an optimization process that minimizes some objective function.
(Other visualization-based explanation methods primarily highlight pixels in the target image that are somehow correlated with a model's decision \cite{simonyanDeepConvolutionalNetworks2014,selvaraju2017grad,ribeiroWhyShouldTrust2016}, but will not be discussed further here).
In many previous applications, this involved pixel-based gradient ascent: starting from white-noise images, and iteratively adjusting the pixels.
This approach has been used to generate equivalence classes (metamers) for image-computable vision models, leading to significant impact in both computer vision and human vision science \cite{balasSummarystatisticRepresentationPeripheral2009, portillaParametricTextureModel,freemanMetamersVentralStream2011,wallisTestingModelsPeripheral2016, boehmUnderstandingCrossModelPerceptual2025, Broderick_2023, waltonBlurRealtimeVentral2021, balasTextureSynthesisPerception2006, rosenholtzSummaryStatisticRepresentation2012, featherModelMetamersReveal2023, featherMetamersNeuralNetworks2019}.
It is also possible to synthesize images that efficiently discriminate between competing models \cite{golanControversialStimuliPitting2020, wang_maximum_2008, berardinoEigenDistortionsHierarchicalRepresentations2017, featherDiscriminatingImageRepresentations2025, boehmUnderstandingCrossModelPerceptual2025}.
\textbf{Feature visualization} and related techniques \cite{erhanVisualizingHigherLayerFeatures2009, mahendranUnderstandingDeepImage2015, olahFeatureVisualization2017, nguyenUnderstandingNeuralNetworks2019} are also forms of synthesis-based explanation.
Here, images are typically generated to maximize the activation of a network layer or node.
These methods are argued to provide qualitative insights into network activations and therefore explanations for their decisions, though the usefulness of these methods in providing humans with unique explanatory information has recently been called into question \cite{zimmermann2021how, borowskiExemplaryNaturalImages2021, geirhosDontTrustYour2024}.
Another approach has been proposed by Veeravasarapu et~al. \cite{veeravasarapuAdversariallyTunedScene2017} by training a GAN using a Bayesian generative model receiving a set of scene parameters and a CNN as a discrimnator to optimize the generative model via Bayesian updates to produce 3D scenes in a probabilistic manner.

To pick examples in which to test our method, we considered both general and more specific image-computable representations.
First, \textbf{general representations (foundation models)} are interesting due to their ability to generalize to other visual tasks with minimal further training.
Here we use \textit{CLIP} \cite{radfordLearningTransferableVisual2021} -- a multi-modal embedding model aligning textual and visual information -- as an example general representation, due to its computational tractability and wide use as a backbone for multi-modal and visual tasks \cite{poole2022dreamfusion, mohammad2022clip, michel2022text2mesh, sanghi2022clip}.
We additionally probe \textit{DINOv2} \cite{caronEmergingPropertiesSelfSupervised2021, oquab2024dinov2learningrobustvisual}, which we further refer to as \textit{DINO}, a self-supervised vision transformer whose dense features are widely used as a general-purpose visual backbone, providing a second foundation-model representation that is trained without language supervision.
Second, because some of our evaluations are qualitative and we are interested in perceptual similarity, we also consider \textbf{perceptual similarity metrics}, which are created to try to measure the similarity or discriminability of image pairs for humans (for recent overviews see \cite{kastryulinPyTorchImageQuality2022, zhuHowEvaluateSemantic2024}).
We use \textit{LPIPS} \cite{zhangUnreasonableEffectivenessDeep2018} as a popular example, which uses a \textit{VGG} backbone and its activations to measure perceptual similarity, with fine-tuning on human similarity judgments. 
We also test the standalone \textit{VGG} \cite{simonyan2014very}, the \textit{CNN} backbone of \textit{LPIPS}.
Third, previous work identified that common neural networks exhibit a bias towards using texture information to categorize images, whereas humans tended to preferentially categorize by shape (\textbf{texture vs shape bias}; \cite{geirhosIMAGENETTRAINEDCNNSARE2019}).
This work further showed that training on \textit{stylized ImageNet (SIN)}, in which texture information was decorrelated with shape information via style transfer, increased shape-bias in these networks.
We therefore compare the shape reconstruction performance of \textit{ImageNet}-trained and SIN-trained versions of \text{ResNet50} (\textit{ResNet-50-SIN}), with the hypothesis that the shape-biased network should perform better at 3D shape reconstruction due to the more disentangled representation of shape and material presumably afforded by SIN-training.

\textbf{Physically based rendering}
is an area of computer graphics that seeks to solve physically based light transport often including light-surface or light-volume interactions formulated by the \textit{rendering equation} \cite{kajiyaRENDERINGEQUATION1986}.
The rendering equation is a recursive formulation, which results in a high-dimensional integration problem with no analytic solution. 
Approximating the equation requires Monte Carlo methods alongside physical models accounting for energy conservation, i.e., the strength of the light falls off the farther it travels and the number of times it reflects or refracts from a surface.
\textbf{Differentiable rendering} 
allows the optimization of the physical properties of a scene by differentiating the Monte Carlo integration in the rendering equation.
The first approaches differentiating the rasterization pipeline appeared in 2014 \cite{loperOpenDRApproximateDifferentiable2014}, allowing the reconstruction of scene parameters such as an object's shape and material.
Third-generation differentiable renderers include \textit{Mitsuba3} \cite{Mitsuba3}, \textit{PyTorch3D} \cite{raviAccelerating3DDeep2020}, \textit{nvdiffrast} \cite{laineModularPrimitivesHighPerformance2020}, and \textit{psdr} \cite{zhangPathspaceDifferentiableRendering2020}.
However, unlike \textit{PyTorch3D} and \textit{nvdiffrast}, \textit{Mitsuba3} and \textit{psdr} can differentiate through integrators solving physically based rendering (path tracing) while preserving derivatives of the operations, allowing the use of gradient-based optimization methods in PBDR.
Actually getting general physically based differentiable rendering to work has a number of technical challenges, caused, for example, by discontinuities with changes in lighting or due to edges, which can interfere with or bias gradient computation.
Mitigating these discontinuities and making PBDR computationally tractable are active research areas in the field of differentiable rendering \cite{nimier-davidRadiativeBackpropagationAdjoint2020, viciniPathReplayBackpropagation2021, nicoletLargeStepsInverse2021, zhangProjectiveSamplingDifferentiable2023, zhangManyWorldsInverseRendering2024, jakobDrJitJustInTimeCompiler}; we build upon these advances here.

\subsection{Preliminaries}
In this section, we introduce the preliminaries for our approach to reconstructing model metamers via differentiable rendering.
We first define model metamers, and then recap the physically based (differentiable) rendering machinery that MRD builds on. Readers already familiar with physically based differentiable rendering may skip the rendering preliminaries and continue to the Methods (Section~\ref{sec:general-methods}); the model-metamer definition below, however, is central to how we interpret our results.

Our goal is to reconstruct a scene and its parameters (or a subset), given a set of latent representations of ground-truth images.
Due to the nature of stochastic gradient descent, the reconstruction will approximate the ground truth, and therefore, will inevitably lead to a different set of scene parameters that are perceived as visually similar by the neural networks used. 
For the reconstruction, we use six different neural networks, which we categorize into classical convolutional neural networks, perceptual metrics, and, lastly, a pair of modern vision transformers.
In the following, we will briefly recap the methods from the different fields on which our work is based.

\paragraph{Model Metamers} We define model metamers as physically-different stimuli that produce the same latent representation, because latent equivalence is the most direct and principled notion of representational indistinguishability in deep networks. Modern models operate entirely on their internal feature representations, and any downstream behavior is fully determined by these latent codes. Inputs that map to the same point (or an equivalence neighborhood) in latent space, therefore, belong to the same representational equivalence class and are functionally indistinguishable to the model. This definition also captures the model’s learned invariances, respects the geometry of its representation, and avoids the limitations of pixel-based or perceptual similarity metrics.

Operationalizing this definition requires a reference level for ``the same'' latent representation, because exact latent equality is never reached under stochastic, physically constrained optimization. We obtain this reference from a baseline that is optimized directly against the ground-truth scene in image space (Section~\ref{sec:results}, and note that the baseline itself does not use the network in its loss). The baseline tells us how similar the target and an image-optimized scene are for the model (measured in its latent space). 
We then declare a reconstruction (optimized against the model's latent directly) a metamer when its latent similarity to the target reaches or exceeds this baseline level. Because a single scalar can hide important structure, we assess this similarity with several complementary metrics (Section~\ref{sec:results}) rather than one, and treat their agreement or disagreement as diagnostic.

\paragraph{Physically Based Rendering}
Physically based rendering models image formation as the evaluation of a light transport operator parameterized by a set of scene parameters~$\pi$, where 
\begin{align*}
    \mathcal{S} := \{\pi | \pi \in \mathbb{R}^{n_\pi}, n_\pi \in \mathbb{N} \}.
\end{align*}
Here, we summarize the key existing PBDR methods for readers unfamiliar with this literature.
We denote the rendering function by
\begin{align}
    f(\mathcal{S}) : \bigcup_{n \in \mathbb{N}} \mathbb{R}^{n} \rightarrow \mathbb{R}^{3 \times \mathrm{width} \times \mathrm{height}},
\end{align}
where the output is a color image produced by simulating the propagation of light through the scene.
\paragraph{The Rendering Equation.}
Image formation is governed by the \emph{rendering equation}~\cite{kajiyaRENDERINGEQUATION1986}, which recursively expresses the outgoing radiance~$L_o$ at a surface point~$\mathbf{x}$ in direction~$\omega_o$ as
\begin{align}
    L_o(\mathbf{x},\omega_o)
    &= L_e(\mathbf{x},\omega_o) 
     + \int_{S^2}
        f_s(\mathbf{x}, \omega_i, \omega_o)\,
        L_i(\mathbf{x}, \omega_i)\,
        \cos\theta_i \,
        \mathrm{d}\omega_i ,
    \label{eq:re_surface}
\end{align}
where $L_e$ denotes emitted radiance and $L_i$ the incident radiance arriving from direction $\omega_i$ over the hemisphere $S^2$.  
The BRDF $f_s$ describes the directional redistribution of incident irradiance into outgoing radiance.  
The cosine factor $\cos\theta_i = \langle \omega_i, \mathbf{n} \rangle$ accounts for geometric foreshortening with respect to the surface normal $\mathbf{n}$.

Because the rendering equation is recursive, $L_i$ generally depends on further scattering events along the path traced by the incoming ray.  
Analytic solutions are therefore infeasible for realistic scenes.

\paragraph{Monte Carlo Approximation.}
To evaluate the integral in Equation~\eqref{eq:re_surface}, we use Monte Carlo (MC) integration:
\begin{align}
    L_o(\mathbf{x},\omega_o) &= L_e(\mathbf{x},\omega_o) + \int_{S^2} f_s(\mathbf{x},\omega_i,\omega_o)\, L_i(\mathbf{x},\omega_i)\, \cos\theta_i \, \mathrm{d}\omega_i \\
 &\approx L_e(\mathbf{x},\omega_o) + \frac{1}{N}\sum_{i=1}^{N} \frac{ f_s(\mathbf{x}, X_i, \omega_o)\, L_i(\mathbf{x}, X_i)\, \cos\theta_i }{ p(X_i) },
\end{align}
where $X_i$ are samples drawn from a distribution with density $p$.  
Variance reduction strategies such as multiple importance sampling and stratified sampling improve estimator robustness~\cite{veachRobustMonteCarlo}.

\paragraph{Path-Space Formulation.}

\begin{figure}
    \includegraphics[width=0.45\textwidth]{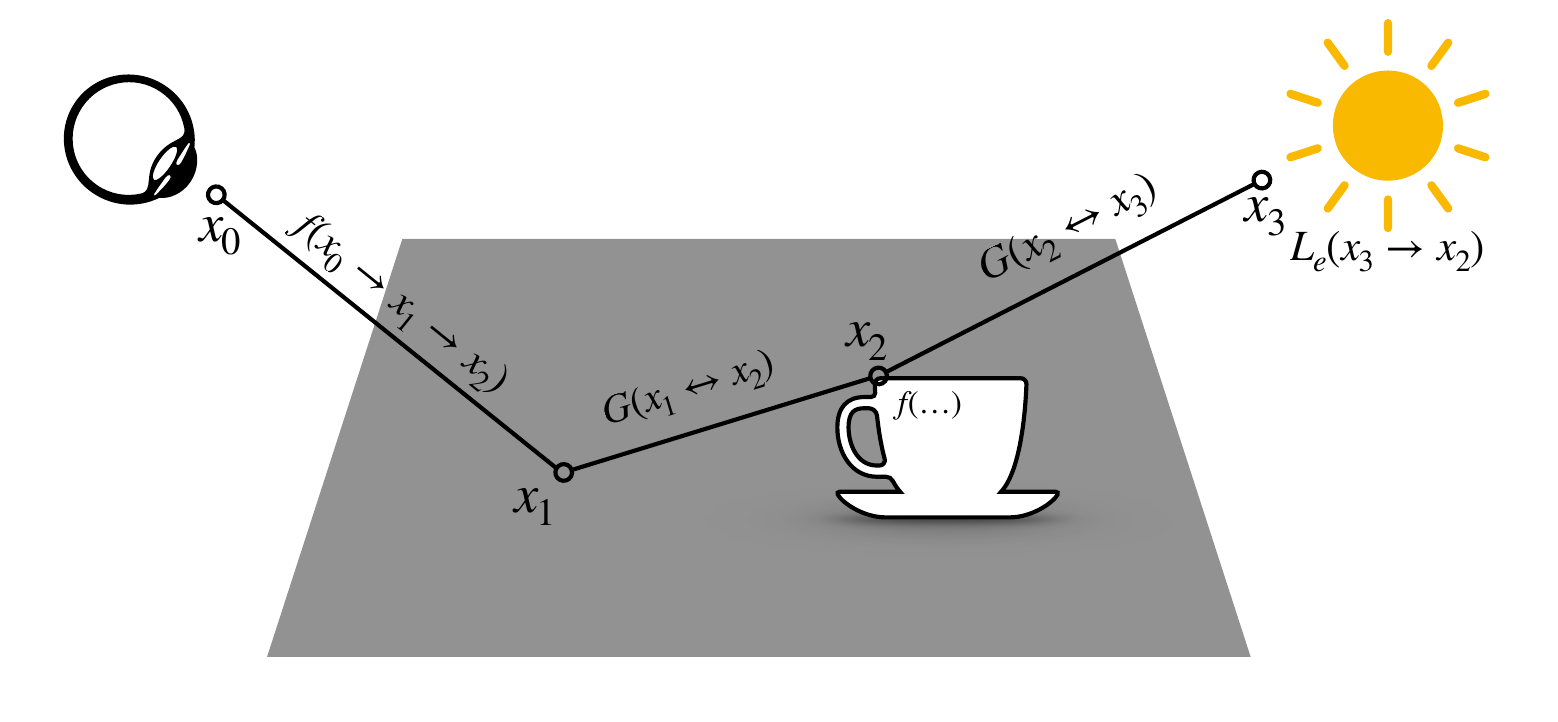}
    \caption{An example light path starting at \( x_0 \) -- the sensor -- computing the BSDF at \( x_1 \) as \( f(x_0 \rightarrow x_{1} \rightarrow x_{2}) \). We also check if there is any occlusion between \( G(x_{1} \leftrightarrow x_{2})\). We stop tracing if: (1) we hit an emitter, (2) terminate because the number of bounces does not meaningfully contribute to the final value, (3) the path between two vertices is degenerate because it is occluded by another object, and lastly (4) the path is not intersecting anything (sky).}\label{fig:lightpath}
\end{figure}

A more general formulation of light transport is obtained by rewriting the rendering equation in the space of all possible light paths~$\Omega$ rather than per-surface interactions.  
Following Veach~\cite{veachRobustMonteCarlo}, the contribution of pixel~$j$ can be written as
\begin{align}
    I_j
    &= \int_{\Omega}
         f_j(\bar{\mathbf{x}})\,
         \mathrm{d}\mu(\bar{\mathbf{x}}),
    \label{eq:path_int}
\end{align}
where $\bar{\mathbf{x}} = \mathbf{x}_0\mathbf{x}_1\mathbf{x}_2\cdots\mathbf{x}_k$ is a path beginning at a point $\mathbf{x}_0$ on the sensor, undergoing a sequence of scattering events, and terminating at an emitter's location.  
The path contribution function decomposes as
\begin{align}
    f_j(\bar{\mathbf{x}})
    = 
    \underbrace{
        G(\mathbf{x}_0 \leftrightarrow \mathbf{x}_1)
    }_{ \text{geometry + visibility} }
    \;
    \underbrace{
        f_s(\mathbf{x}_0 \rightarrow \mathbf{x}_1 \rightarrow \mathbf{x}_2)
    }_{ \text{material} }
    \;\cdots
    \underbrace{
        L_e(\mathbf{x}_k \rightarrow \mathbf{x}_{k-1})
    }_{ \text{emission} },
\end{align}
where the geometry term $G$ includes inverse-square falloff, foreshortening, and visibility via a Dirac-delta that nulls contributions from occluded configurations.  
The measure $\mu$ encodes the sampling distribution over path space. We show an example path in Figure \ref{fig:lightpath}.

\paragraph{Inverse Rendering via Differentiable Light Transport.}
Our aim is to estimate scene parameters~$\pi$ from observed images~$y$ by minimizing a reconstruction loss between $f(\pi)$ and $y$.  
This requires backpropagating gradients through the light transport simulation, i.e.,
\begin{align}
    \pi^\star = \arg\min_{\pi}\; \mathcal{L}(f(\pi), y),
\end{align}
necessitating the derivatives of the rendering equation with respect to~$\pi$.  
Since the outgoing radiance $L_o$ depends on scene parameters through geometry, materials, lighting, and visibility, we differentiate Eq.~\eqref{eq:re_surface} to obtain
\begin{align}
    \frac{\mathrm{d} L_o}{\mathrm{d}\pi}
    &= 
    \frac{\mathrm{d} L_e}{\mathrm{d}\pi}
    + 
    \frac{\mathrm{d}}{\mathrm{d}\pi}
    \int_{S^2}
        f_s(\mathbf{x}, \omega_i, \omega_o)\,
        L_i(\mathbf{x}, \omega_i)\,
        \cos\theta_i
    \,\mathrm{d}\omega_i .
    \label{eq:re_grad}
\end{align}
The derivative includes contributions from (1) the BRDF parameters, (2) geometric changes affecting $\mathbf{x}$, normal directions, and thus $\cos\theta_i$, (3) derivatives of incident radiance $L_i$, and (4) visibility terms, which introduce discontinuities that must be handled via techniques such as soft visibility, edge sampling, or regularization~\cite{Mitsuba3, viciniPathReplayBackpropagation2021}.
Equation~\eqref{eq:re_grad} forms the foundation physically based differentiable rendering.  
It enables gradient-based optimization over complex, physically grounded image formation models and underlies the inverse-rendering procedure used in our experiments.

\paragraph{Handling Visibility Discontinuities.}
Differentiating light transport is complicated by visibility discontinuities, which occur at object boundaries and shadow edges.  
Formally, the derivative of the image with respect to scene parameters~$\pi$ decomposes into two components:
  (1) \textit{Interior derivatives}, arising from smooth variations of geometry, materials, and illumination along paths whose visibility does not change.
  (2) \textit{Boundary derivatives}, arising from changes in ray visibility when geometry crosses occlusion boundaries. These terms dominate near silhouettes and shadow boundaries and are responsible for discontinuous behavior in the rendering function.
While standard path‐wise differentiation techniques handle interior terms, boundary terms require specialized sampling near the visibility manifold.  
Recent work by Zhang et~al.~\cite{zhangProjectiveSamplingDifferentiable2023} provides a principled treatment of these boundary contributions by explicitly sampling near geometry–visibility boundaries. Their approach yields unbiased estimates of both interior and boundary terms and substantially improves gradient accuracy for inverse rendering tasks.

Following the shape-derivative analysis of Zhang et~al.~\cite{zhangProjectiveSamplingDifferentiable2023},
the derivative of a pixel intensity $I_j$ with respect to scene parameters $\pi$
decomposes into an \emph{interior} term and a \emph{boundary} term:
\begin{align}
    \frac{d I_j}{d\pi}
    = 
    \underbrace{
        \int_{\Omega}
            \frac{\partial f_j(\bar{\mathbf{x}})}{\partial \pi}
        \, d\mu(\bar{\mathbf{x}})
    }_{\text{interior term}}
    +
    \underbrace{
        \int_{\partial\Omega(\pi)}
            f_j(\bar{\mathbf{x}})
            \, (\mathbf{v} \cdot \mathbf{n})
        \, d\sigma(\bar{\mathbf{x}})
    }_{\text{boundary term}},
    \label{eq:interior_boundary_decomp}
\end{align}
where $\Omega$ is the set of valid light paths, $\partial\Omega(\pi)$ denotes the geometry visibility boundary (e.g., silhouettes or shadow edges), $\mathbf{n}$ is the boundary normal, and $\mathbf{v}$ is the velocity field induced by the change of scene parameters. The interior term corresponds to smooth variations of geometry, materials, or lighting along paths whose visibility remains unchanged. The boundary term accounts for discontinuities introduced when rays cross visibility boundaries. Zhang et~al.\ show that accurate gradient estimation requires explicitly sampling near $\partial\Omega(\pi)$, yielding unbiased estimates of both terms and significantly improving inverse-rendering stability.

\begin{table*}[h]
    \centering
    \caption{Optimization target per network. For each network we list the feature(s) used as the latent target and the loss that is minimized between the target-scene and reconstruction-scene latents during optimization. For the embedding networks (ResNet-50, ResNet-50-SIN, CLIP, DINO) the objective is the mean squared error (MSE) on the latent; for VGG and LPIPS we use their native multi-layer perceptual distance (LPIPS additionally applies a learned, human-aligned weighting on top of the VGG features). This is the loss referred to throughout as the perceptual/latent objective.}
    \label{tab:layers}
    \begin{tabular}{lccc}
        \toprule
        \textbf{Neural Network} & \textbf{Layer / feature} & \textbf{Optimization loss} & \textbf{Remarks} \\
        \midrule
        ResNet-50 & 2048 & MSE on latent & Penultimate, no activation \\
        ResNet-50-SIN & 2048 & MSE on latent & Penultimate, no activation\\
        \midrule
        CLIP & \( \mathbb{R}^{260 \times 768} \) & MSE on latent & vision encoder out, no activation, dense features \\
        DINO & \( \mathbb{R}^{260 \times 768} \) & MSE on latent & no activation, dense features \\
        \midrule
        VGG & 5 & multi-layer \( \ell_2 \) (perceptual) & spatially averaged \( \ell_2 \) distances\\
        LPIPS (VGG) & 1 & learned-weighted \( \ell_2 \) (LPIPS) & same as VGG, but with similarity network on top\\
        \bottomrule
    \end{tabular}
\end{table*}
\paragraph{Neural Networks}\label{sec:nn}
In this work, we probe six neural networks to examine how sensitive their internal representations are to shape and material. All neural networks run inference and do not update their weights, i.e., model weights are frozen.

\noindent{} (1) Convolutional Neural Networks. We begin with two convolutional architectures: \textit{ResNet50} and \textit{ResNet50-SIN} \cite{krizhevsky2012imagenet, geirhosIMAGENETTRAINEDCNNSARE2019}, where the latter is trained on \textit{Stylized ImageNet} to increase shape bias. The \textit{ResNet} architecture relies on residual connections that facilitate gradient flow in deep networks, thereby helping earlier layers avoid the effects of vanishing gradients. These models are widely used for object recognition and as generic feature backbones. For our purposes, we use the ImageNet-1k weights for \textit{ResNet50} and the Stylized ImageNet weights for \textit{ResNet50-SIN}. In both cases, we discard the final classification layer and use the 2048-dimensional penultimate layer as our target latent space. Although earlier layers retain more photometric detail \cite{gatys2015neuralalgorithmartisticstyle, featherModelMetamersReveal2023, featherMetamersNeuralNetworks2019}, our aim is not necessarily to reconstruct the original image, but to reconstruct the conceptual representation encoded in the network’s latent space (we return to this point in the Discussion).

\noindent{} (2) Perceptual Metrics. A second pair of models is drawn from networks used for perceptual similarity metrics. We evaluate \textit{LPIPS} (with a \textit{VGG} backbone) \cite{zhangUnreasonableEffectivenessDeep2018} and compare it to the same \textit{VGG} architecture without the \textit{LPIPS} perceptual-alignment training. 
Both metrics aggregate multi-layer \textit{VGG} features to compute similarity scores, while \textit{LPIPS} further learns human-aligned weighting from human similarity judgments of image patches.

\noindent{} (3) Vision Transformers. Finally, we include two recent Vision Transformer (ViT) models -- \textit{DINO} \cite{caronEmergingPropertiesSelfSupervised2021, oquab2024dinov2learningrobustvisual} and \textit{CLIP} \cite{radfordLearningTransferableVisual2021}. These models differ from CNNs by relying primarily on self-attention mechanisms, performing scaled dot-product attention over image patches to build global representations. \textit{CLIP} learns an aligned representation between modalities (language and visual features).

In Table~\ref{tab:layers}, we provide an overview of the layers used for the optimization.

\paragraph{Scene}\label{sec:assets}
Because we are using Mitsuba \cite{Mitsuba3}, we need to define our 3D scene in a Python dictionary or use the XML scene description. In contrast, neural networks do not require an initial scene configuration; we can learn the scene's properties directly from the images. Therefore, with Mitsuba, we need to initialize a scene that we adjust with gradient updates. In this section, we will introduce the assets and the scene setup.

In this work we use artificial scenes in order to have a well-defined ground truth.
We use two separate scenes for shape and material reconstruction.

\paragraph{Shape Reconstruction} The scene consists of multiple environment maps that account for different lighting conditions, and a single floating object, with a diffuse, non-textured material positioned in the center of the scene. 
We use a floating object for shape reconstructions in order to allow unobstructed views from all camera perspectives. 

\paragraph{Material Reconstruction} The scene for the material reconstruction differs from the shape reconstruction. We place an object on a checkerboard and light the scene with a single environment map. The object will start from an initial, textured Principled BSDF (\textit{Disney BSDF}) \cite{burleyPhysicallyBasedShading,burley2015extending}.

\paragraph{Camera Views} The reconstruction of scene parameters, such as shape and reflectance, benefits from multiple views, which constrain the optimization process. For shape reconstruction, we sample \( n- \)camera origins using the Fibonacci lattice on a unit sphere, and for material reconstruction, we view the object from front/back and both sides. From these camera positions, we render our \( n- \)ground truth images and apply the \textit{Optix Denoiser} \cite{chaitanyaInteractiveReconstructionMonte2017} available in \textit{Mitsuba}.


\paragraph{Image Format} To ensure that the rendered images produced by Mitsuba 3 are compatible with neural networks, we convert the output from the renderer’s default high-dynamic-range (HDR) linear radiance representation into a standard display-referred sRGB space. Since the raw Mitsuba outputs contain unbounded scene-referred radiance values that are out-of-distribution for networks such as \textit{ResNet}, we first transform the renderer’s internal RGB basis to linear sRGB, and then apply Reinhard tonemapping to compress the dynamic range into a bounded interval. Reinhard tonemapping compresses unbounded HDR radiance into a bounded, perceptually plausible range by applying a simple luminance normalization that preserves detail while preventing saturation. Finally, we apply the sRGB transfer function (gamma encoding), yielding images with normalized, perceptually uniform intensities suitable for training. This conversion ensures consistent color representation and prevents numerical instabilities arising from unbounded HDR values during learning.

\section{Methods}\label{sec:general-methods}

\paragraph{What is optimized, and against what objective} Before the experiment-specific details, we state the optimization target explicitly, since it is shared across all experiments. In every experiment we render $n$ views of the current scene with the differentiable renderer $f(\pi)$, pass each rendered view through a frozen network $\mathrm{NN}(\cdot)$, and update the scene parameters $\pi$ to minimize the distance between the network's latents for the current scene and for the target scene (the loop in Listing~\ref{lst:base}). Which parameters are optimized depends on the experiment: for \textbf{material} reconstruction we optimize the \textit{Principled BSDF} parameters together with the albedo and (where present) anisotropy textures (Section~\ref{sec:mat-rec}); for \textbf{shape} reconstruction we optimize the vertex positions of a proxy mesh (Section~\ref{sec:shape-rec}). Which latent and which loss is used depends on the network and is summarized in Table~\ref{tab:layers}: for the embedding networks (ResNet-50, ResNet-50-SIN, CLIP, DINO) we minimize the mean squared error on the latent shown in Table~\ref{tab:layers}, whereas for VGG and LPIPS we minimize their native multi-layer perceptual distance. The baseline reconstructions (Section~\ref{sec:results}) optimize the same scene parameters but against an image-space loss instead of a network latent, and are used only to calibrate the metamer criterion.

We implement all of our experiments using Python 3.12, \textit{PyTorch} \cite{NEURIPS2019_9015}, and \textit{Mitsuba 3} \cite{Mitsuba3}, which is a suitable PBDR for our experiments, and mitigates many of the discontinuity issues mentioned above (\cite{nimier-davidRadiativeBackpropagationAdjoint2020, viciniPathReplayBackpropagation2021, nicoletLargeStepsInverse2021, zhangProjectiveSamplingDifferentiable2023, zhangManyWorldsInverseRendering2024}). For all of our experiments, we also define an early stopping criterion: the experiment stops if the loss does not improve for 50 epochs. We run all of our shape reconstruction experiments on a workstation with a single RTX 4090 with 24 GB of VRAM using Ubuntu 24.04.5 LTS. For our material reconstruction experiments, more VRAM is required due to the path tracing integrator; for these experiments, we use an RTX 6000 with 48 GB of VRAM running on an identical system environment. We use pretrained models and implementations of these if available \cite{geirhosIMAGENETTRAINEDCNNSARE2019, radfordLearningTransferableVisual2021}. Our code and plots for all experiments will be available here: \url{https://doi.org/10.5281/zenodo.21374307}.

\begin{figure}
    \includegraphics[width=\linewidth]{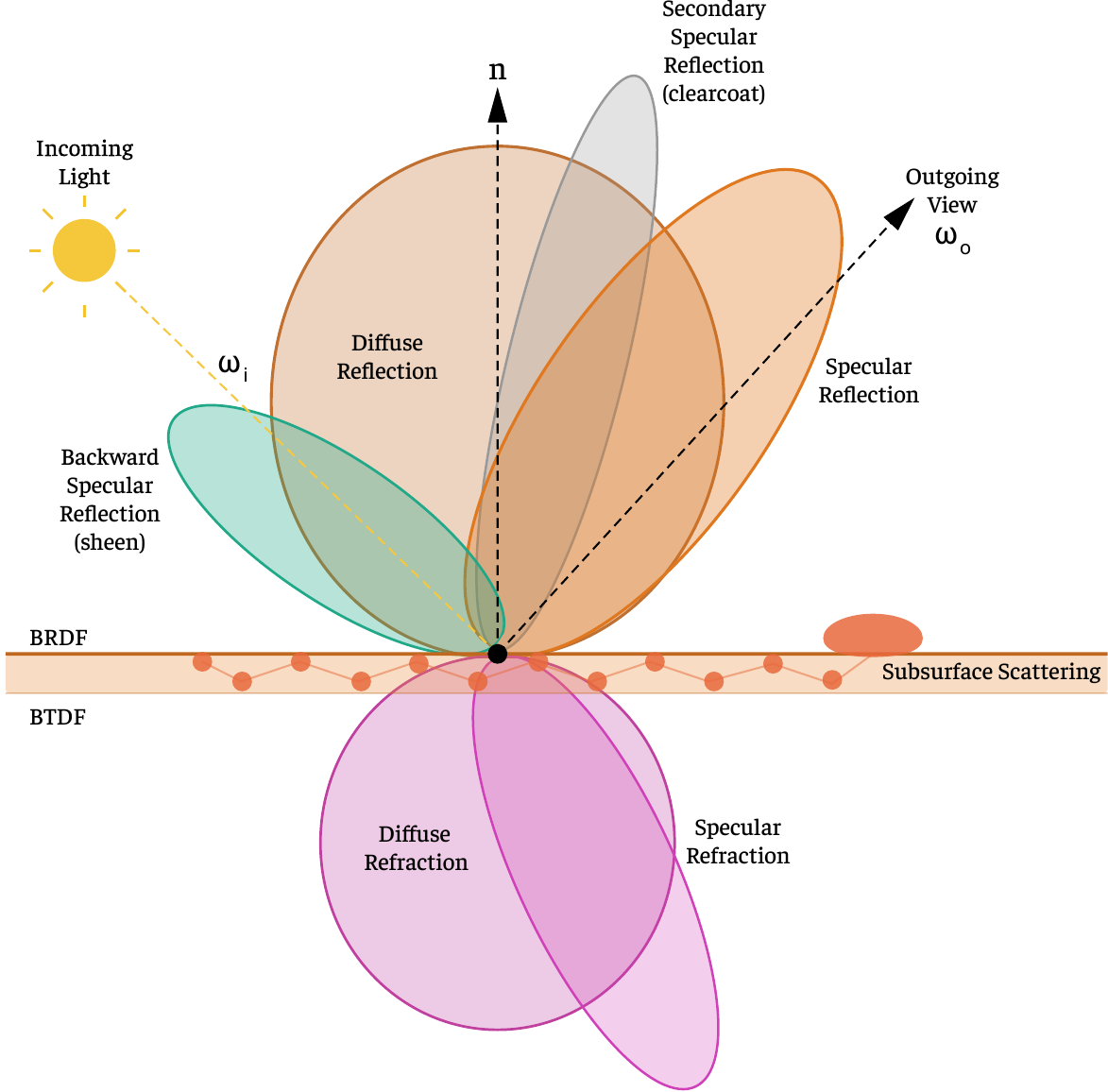}
    \caption{Principled BSDF. The BSDF contains multiple lobes that visualize how light reflects/refracts. The surface roughness will influence how wide these lobes are, thus, how sharp the reflections are.}\label{fig:bsdf}
\end{figure}

\subsection{Material Reconstruction}\label{sec:mat-rec}
Every object in our scene has one or multiple materials attached to it, that define how light interacts with the surface of an object. It is one of the main drivers of how light is transported through space. 

In our material optimization experiments, we make parameters of the BSDF and texture available as targets for optimization, fixing other scene parameters (geometry, lighting, etc). Our optimization then aims to reconstruct the material. The difficulty in this task lies in the absence of a one-to-one parameter mapping: the target materials are not themselves \textit{Principled BSDF}s, so there is, in general, no single setting of the \textit{Principled BSDF} parameters that reproduces the target exactly. However, all materials we select as our target can be modeled by the Principle BSDF. Some materials, such as the Brushed Metal and Aurora, are recorded with a goniophotometer \cite{dupuyAdaptiveParameterizationEfficient2018},  thus have no ground truth parametric Principled BSDF. At a high level, this means the material reconstruction is \textbf{metameric by definition} --- many different parameter settings can produce an essentially indistinguishable appearance, so the optimizer is free to (and generally will) settle on a parameterization that matches the model's representation of the target material without recovering the target's underlying physical parameters, while additionally reconstructing the textures for the albedo and anisotropy.

\paragraph{Principled BSDF} The \textit{Principled BSDF} \cite{burleyPhysicallyBasedShading, burley2015extending} was introduced to provide a production-friendly, art-directable surface model that remains physically plausible while exposing an intuitive parameter set. Rather than representing a single physical model, the \textit{Principled BSDF} blends a small set of carefully chosen analytic lobes -- diffuse, retro-reflective, specular microfacet, clearcoat, sheen, and transmission -- into a unified framework with consistent energy conservation. Each lobe is designed to behave reasonably across the full parameter domain, enabling smooth transitions between dielectric and metallic appearances via a single \textit{metalness} control while preserving reciprocity and approximate Fresnel behavior (which also benefits the optimization due to predictable changes).

\paragraph{Integrator} Reconstructing a \textit{Principled BSDF} in Mitsuba requires the use of a path tracing integrator because the appearance of such materials is determined by a combination of diffuse, specular, microfacet, subsurface, and transmission components that interact nonlinearly with the full distribution of incoming light. Translucent effects cannot be captured using compute-efficient integrators such as \textit{Path Replay Backpropagation} \cite{viciniPathReplayBackpropagation2021}. Path tracing provides an unbiased estimate of the rendering equation, allowing gradients to flow through all physically relevant light transport paths. This is essential for recovering material parameters whose influence may only emerge through indirect bounces, grazing-angle reflections, or jointly through coupled terms such as roughness, index of refraction, and anisotropy. Consequently, a physically based global illumination integrator like path tracing is necessary to ensure that the optimization receives accurate, informative feedback and can faithfully reconstruct the full behavior of a \textit{Principled BSDF}. Further, the path tracing integrator requires more samples per pixel to reconstruct the textures used to parameterize the material.

\paragraph{Parameter Clipping} To ensure physically plausible behavior during optimization, the parameters of the \textit{Principled BSDF} in Mitsuba must be clamped to their valid ranges. Many of these parameters, such as roughness, specular transmission, metalness, and anisotropy, are defined only over restricted intervals $[ 0, 1 ]$, and allowing the optimizer to explore values outside these domains can lead to numerical instabilities, non-physical reflectance, or invalid configurations within the renderer. Clamping prevents such degeneracies by projecting intermediate updates back into the feasible set, ensuring that the material model remains consistent with its underlying physical assumptions throughout the optimization process. This constraint is particularly important when gradients are noisy, as is common in path tracing, where unconstrained parameter updates may otherwise accumulate error and destabilize convergence. By enforcing these bounds, we maintain stable optimization dynamics and guarantee that the reconstructed material remains compliant with the \textit{Principled BSDF} specification.

\begin{figure}
    \includegraphics[width=\linewidth]{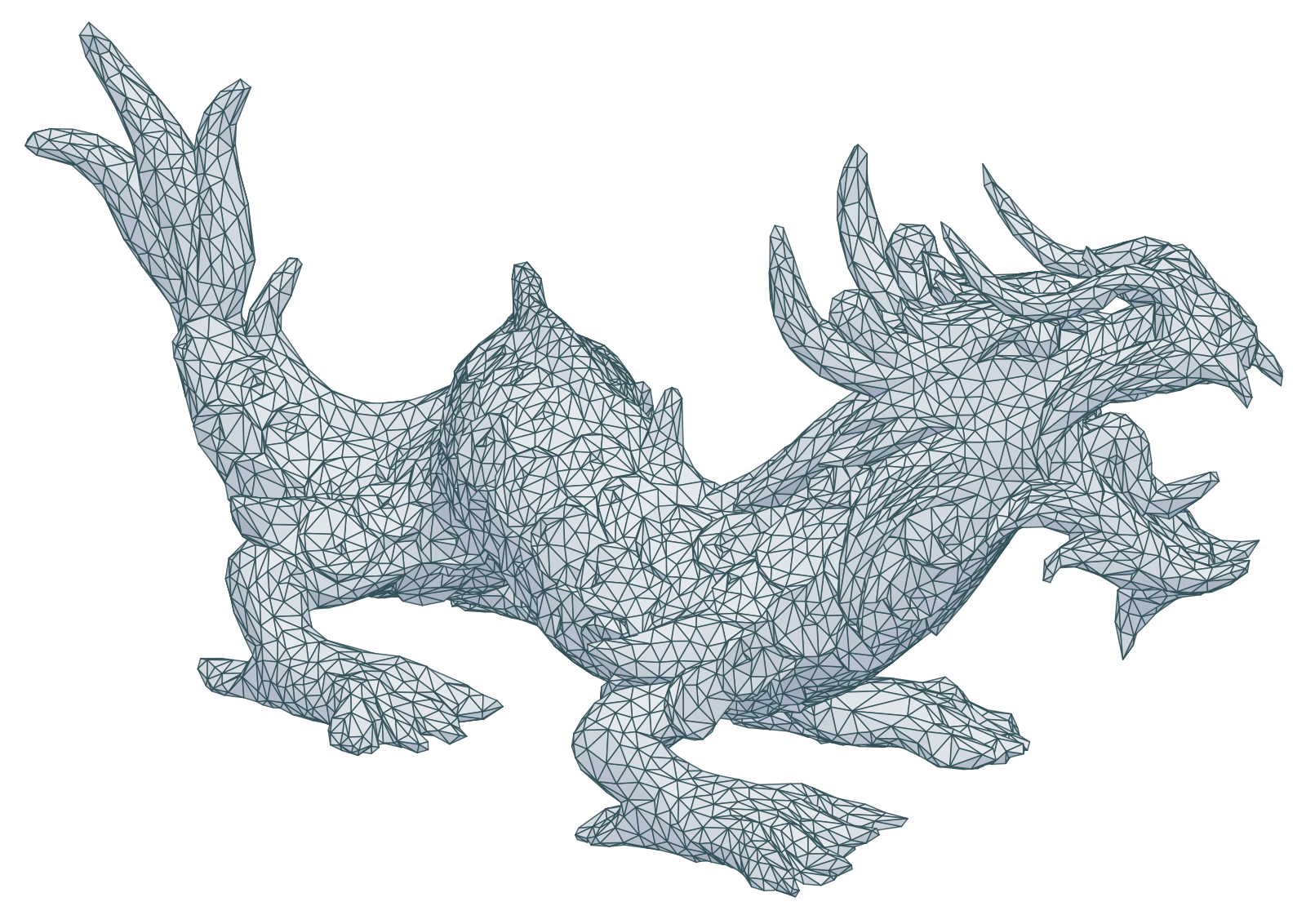}
    \caption{An example mesh of our Dragon model. A triangulated mesh contains a number of vertices. Three vertices form one face. Moving the vertices in space will affect the shape of the face, therefore, deforming the mesh.}\label{fig:mesh}
\end{figure}

\subsection{Shape Reconstruction}\label{sec:shape-rec}
Geometry in computer graphics is often represented as a triangulated mesh using vertices and edges to form faces (Figure \ref{fig:mesh}), and therefore, the surface of an object. A defined set of vertices defines a single face; however, the vertices can still move in the three-dimensional space and form wider or narrower faces/triangles. In the shape reconstruction experiments, we allow moving the vertices in space to model the target shape visible in the reference images. In our experiments, we start from a proxy mesh -- an icosphere with 16000 vertices -- optimizing it towards the target mesh.

\paragraph{Large Steps} Our implementation for the shape reconstruction follows the \textit{Large Steps} method of Nicolet et~al. \cite{nicoletLargeStepsInverse2021}. This method achieves fast and efficient convergence by preconditioning the bias of the gradient steps towards smooth solutions without enforcing smoothness in the final result.
\( \lambda  \) is a new hyperparameter introduced in \cite{nicoletLargeStepsInverse2021} that controls the strength of regularization introduced by the Laplacian matrix. 
Nicolet et al also introduce changes in the Adam \cite{kingmaAdamMethodStochastic2017} optimizer to precondition the gradient update and allow for higher learning rates. 
In our implementation, we use \textit{Mitsuba 3} instead of \textit{nvdiffrast} and \textit{PyTorch}, necessitating adjustment of \( \lambda  \) and the step size, thus deviating from the suggestions in Nicolet et al. \cite{nicoletLargeStepsInverse2021}.

\paragraph{Integrator} 
We use the \textit{direct projective path replay backpropagation} method introduced by Zhang et~al.~\cite{zhangProjectiveSamplingDifferentiable2023} for our shape reconstruction experiments. This integrator is specifically designed to handle the discontinuities that arise when object boundaries move during optimization. It mitigates these boundary discontinuities by using \textit{projective sampling}, where samples generated during the forward pass are reprojected onto nearby geometry boundaries to estimate the boundary derivative stably and with reduced variance. The key contribution of Zhang et~al. is to rewrite the boundary term of the rendering equation into a local integral over projected samples, removing the need to integrate over the full non-local geometry. This projective formulation offers several advantages: (1) it eliminates the complex non-local domain that normally appears in boundary derivatives, (2) it removes the geometric term $G$, which is responsible for much of the variance in classical formulations, and (3) it leads to a significantly simplified expression for common geometries such as smooth closed surfaces and polygonal meshes. A helpful way to build intuition is to consider a simplified 2D example: imagine a curtain in front of a window. Moving the curtain changes the visibility discontinuously. Instead of tracking how the curtain itself moves (which produces the discontinuity), the integrator conceptually ``moves" the image of the window onto the curtain, i.e., it adjusts the integration domain rather than the geometry. This projection makes the derivative well-behaved and substantially reduces estimator variance.

\paragraph{Tesselation} The target mesh can use more vertices than our initial icosphere. If that is the case, we allow our current mesh reconstruction to tessellate its mesh using each side's midpoint \cite{botschRemeshingApproachMultiresolution2004}. We apply this remeshing operation on specific epochs, which we pick after testing our baseline. Tesselating the mesh will always result in an increase in the loss, because we converged to the vertices' position, yet adding more vertices allows the optimization to deal with details on the object's surface. Each tesselation will be followed with a learning rate decay multiplying the previous learning rate by a factor of \( 8e^{-3} \). We apply this decay only after remeshing because the details in the mesh no longer require large changes.

\subsection{Evaluation} 
\label{sec:methods-evaluation}
There is no direct measurement for metamerism in our setting: unlike pixel-space metamer work (e.g. Feather et al.~\cite{featherModelMetamersReveal2023, featherMetamersNeuralNetworks2019}), which matches a network's activations by optimizing the image pixels directly, MRD optimizes physical scene parameters, so there is no a~priori ``correct'' metameric image to compare against. We therefore calibrate against a baseline reconstruction (Section~\ref{sec:baseline}) that drives the scene as close as possible to the ground truth in image space --- crucially without using any network in its loss --- and read out the resulting network latent similarity. This tells us how similar two scenes' latents become under a near-ideal image-space reconstruction, and gives us a principled threshold.

With this in place, we separate two distinct outcomes that we present separately:
\begin{itemize}
    \item \textbf{Did MRD find a model metamer? (primary outcome)} For each experiment, we count how many of our latent metrics (cosine, Pearson, Spearman, and RSA, where available; see Section \ref{sec:evaluation-metrics}) the reconstruction satisfies relative to the baseline --- this is the \# column of Tables~\ref{tab:results_material} and~\ref{tab:results_shape_metamers}, ranging from 0 to 4. Because the four metrics probe different facets of representational equivalence, we report this full per-criterion breakdown rather than collapsing it to a single pass/fail: meeting at least one criterion is suggestive evidence of metamerism, while meeting all four is strong evidence. We are explicit about this because a permissive ``any one criterion'' rule, taken alone, would overstate the count. This is a statement about MRD and about the network's equivalence class, not about whether the original scene was recovered.
    \item \textbf{Did the optimization recover the ground-truth scene parameters? (secondary outcome)} This is a stricter question --- whether the reconstructed material/shape is actually the target one --- which we read from image-space fidelity (\flip{}) and from visual inspection. Note that \flip{} measures appearance fidelity, which is only a proxy for true parameter recovery: a metamer can render close to the target while resting on different physical parameters. A reconstruction can be a metamer (primary) while clearly failing to recover the ground truth (secondary); when this happens, it tells us the network's equivalence class is broad rather than specific.
\end{itemize}

Keeping these outcomes separate is important. For example, a reconstruction reaching baseline similarity means that MRD finds a metamer, but does not tell us whether that model metamer is an accurate recovery of the true scene (see Figures~\ref{fig:translucent-resnet} and~\ref{fig:translucent-dino}).

\begin{figure*}
    \centering
    \includegraphics[width=\textwidth]{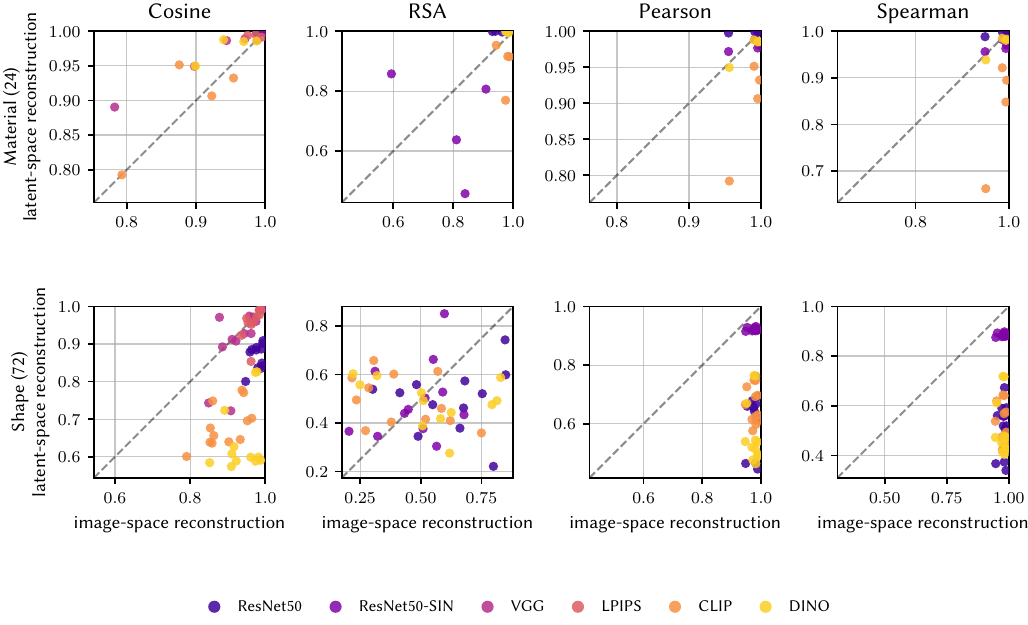}
    \caption{\small Each scatter plot compares a latent-space reconstruction score ($y$-axis) against the corresponding image-space baseline score ($x$-axis) for all material (top, 24 experiments) and shape (bottom, 72 experiments) experiments, across all four similarity scores. Points above the diagonal indicate that the latent-space reconstruction matches or exceeds the baseline --- our criterion for declaring a model metamer. Materials cluster tightly near the top-right corner (most are metamers), whereas shapes are predominantly below the diagonal, and the four metrics dissociate for shapes. We unpack the per-metric counts and the metric dissociations in the text (Sections~\ref{sec:results-material} and~\ref{sec:results-shape}).}
    \label{fig:scatter}
\end{figure*}

\subsection{Baseline experiments}\label{sec:baseline}
\begin{listing}[H]
\begin{lstlisting}
# Requires: PBDR f, parameters pi, loss function L
#           optimizer O, scene S, transformation T

for epoch e in range(1, E+1):
    for sensor v in V:
        pi = T(pi)              # Remesh/clamp
        render_v = f(S, pi, v)  # Render
        l = L(I_v, render_v)    # Compute loss
        l.backward()            # Backpropagation
        O.step()                # Update parameters
        pi.update(O)            # Update scene
\end{lstlisting}
\caption{The default optimization loop used to run the experiments. The only
factor changing is the transformation function \(\mathcal{T}\) applied to
different experiments, such as remeshing for geometry, or clamping values
in the material experiments.}
\label{lst:base}
\end{listing}
Our baseline experiments serve two purposes: (1) they let us check and tune the hyperparameters, and (2) they calibrate the scores, i.e. they record what each score looks like for a reconstruction that is driven directly towards the ground-truth shape/material. 
This calibration is what makes our metamer criterion well-defined. 
The setting is identical to the experiments in Sections~\ref{sec:mat-rec} and~\ref{sec:shape-rec}, with one key difference: the baselines do not use the neural networks in the optimization at all, but compute an error directly in image-space.

The image-space loss differs by experiment type. For \textbf{shape}, the baseline uses \textit{mean-absolute error} between rendered and ground-truth views, which is robust to outliers. For \textbf{material}, a plain pixel error is not adequate: material reconstruction relies on an unbiased path tracing integrator within Mitsuba \cite{Mitsuba3}, whose renders carry Monte Carlo noise, so an $L_2$/MSE loss against a single noisy render yields biased gradients (the squared noise does not average to zero). We therefore use the \textit{Dual-Buffer loss} of Deng et~al. \cite{deng2022reconstructing} for the material baseline only; it has good convergence properties while requiring fewer samples per pixel, making path tracing usable here. The \textit{Dual-Buffer loss} renders the scene twice with the same parameters but different random seeds, giving two independent noisy estimates $\hat{y}, \hat{y}\prime$, and correlates their errors against the ground-truth view $y$ so that the uncorrelated noise cancels in expectation,
\begin{align}
    \mathcal{L}_{dual}(\hat{y}, \hat{y}\prime, y) &= \frac{1}{N} \sum_{i=0}^N \left( (\hat{y} - y) \cdot(\hat{y}\prime - y) \right)^2.
\end{align}
The general optimization loop is displayed in Listing~\ref{lst:base}, and the overview of the hyperparameters for each scene is shown in Table~\ref{tab:hparams}.

To quantify the reconstruction fidelity in the image-space, let $x_i^{(n)}$ denote the reconstruction at step $i$ under the $n$-th view, and let $\hat{x}^{(n)}$ denote the corresponding ground truth. Let $m^{(n)}$ be a binary foreground mask such that $m^{(n)} \odot x$ suppresses all background regions of image $x$. We then define the baseline score as
\begin{equation}
    \text{score}_{\text{baseline}} = \text{score}\!\left(\text{NN}\!\left(m^{(n)} \odot \hat{x}^{(n)}\right),\, \text{NN}\!\left(m^{(n)} \odot x_i^{(n)}\right)\right),
\end{equation}
where $\text{NN}(\cdot)$ denotes the projection of an image into the network's latent space. The mask $m^{(n)}$ removes the environment-map background so that the latent similarity is computed on the object rather than on the (identical) surroundings. We apply this mask only to the shape reconstruction experiments, where the object is lit by direct lighting from the environment map and the background contributes nothing to the shape. We deliberately do not mask the material experiments, because there the object's appearance depends on global illumination and indirect lighting (e.g. bounces from the checkerboard ground onto the object); masking those regions out would discard appearance cues that are part of what defines the material.

\begin{table*}[htbp]
    \centering
    \caption{Hyperparameters for material and shape reconstruction. Most hyperparameters were selected based on prior work~\cite{nicoletLargeStepsInverse2021}. We determine the number of views based on visible surface coverage. Textures have RGB color channels at \(256\times256\) resolution for all materials except the translucent one (\(720\times480\)); the anisotropic texture in the brushed metal experiment uses a single channel at the same resolution. Shape hyperparameters were further adjusted manually to work within a rendering objective. For material reconstruction we optimize the parameters using the Dual Buffer loss, while for shape reconstruction we use the mean absolute error.}
    \label{tab:hparams}
    \begin{tabular}{lcccccc}
        \toprule
        \textbf{} & \textbf{\# Views} & $\boldsymbol{\lambda_{\text{reg}}}$ & \textbf{LR} & \textbf{Epochs} & \textbf{Remesh Epochs} & \textbf{Material Parameters} \\
        \midrule
        \multicolumn{7}{l}{\textit{Material Reconstruction}} \\
        \midrule
        Translucent   & 5  & — & $0.05$            & 500  & — & \small roughness, $\eta$, albedo texture, spec\_trans \\
        Diffuse       & 5  & — & $0.05$            & 500  & — & \small all (albedo texture) \\
        Brushed Metal & 5  & — & $0.05$            & 500  & — & \small all (albedo \& anisotropic texture) \\
        Aurora        & 5  & — & $0.05$            & 500  & — & \small all (albedo texture) \\
        \midrule
        \multicolumn{7}{l}{\textit{Shape Reconstruction}} \\
        \midrule
        Dragon        & 25 & 15 & $1\times10^{-1}$ & 1000 & \small [5, 25, 50, 100, 150, 250, 350, 450] & — \\
        Lion Statue   & 25 & 15 & $1\times10^{-1}$ & 1000 & \small [5, 25, 50, 100, 150, 250, 350, 450] & — \\
        Dog           & 25 & 15 & $1\times10^{-1}$ & 1000 & \small [5, 50, 100, 150, 250, 350, 450]     & — \\
        Suzanne       & 8  & 25 & $1\times10^{-1}$ & 1000 & \small [100, 200, 300, 400]                 & — \\
        \bottomrule
    \end{tabular}
\end{table*}

\subsection{Evaluation Measurements}
\label{sec:evaluation-metrics}
No single similarity captures entirely the idea that two latent representations are the same.
We therefore use four complementary latent-space similarity measurements, each probing a different facet of representational agreement: (1) \textit{unit hypersphere (cosine) similarity}, which measures only the direction of a latent vector; (2) the \textit{Pearson} correlation, which measures whether the individual feature dimensions co-vary linearly between reconstruction and target; (3) the \textit{Spearman} correlation, the rank-based (monotonic) counterpart of Pearson, which is robust to non-linear but order-preserving distortions; and (4) \textit{representational similarity analysis} (RSA) \cite{kriegeskorte2008representational}, which measures whether the relational geometry across an entire set of views is preserved. These probe progressively stricter notions of equivalence (direction $\rightarrow$ per-dimension structure $\rightarrow$ inter-view geometry), and their disagreements are themselves informative: for instance, a high cosine similarity with a low Pearson correlation indicates that the aggregate direction matches while the internal coordinate structure does not. 

For each score, we then define a binary criterion: if the score is higher for the reconstruction using a model's latent space than for the image-space baseline, we count this as evidence that the experiment is a model metamer (the primary outcome of Section~\ref{sec:methods-evaluation}). We further quantify the relative evidence with a Bayes factor for each per-view score. Finally, separate from these latent-space metamer scores, we report the image-space error using \flip{}, which speaks to the secondary outcome --- how well the reconstruction recovers the ground-truth appearance.

\paragraph{Cosine Similarity on the Unit Hypersphere.}
Using the similarity on the unit hypersphere allows us to factor out variations in vector magnitude and focus exclusively on directional information in the latent representations. This is particularly beneficial in our setting, since many latent spaces, especially those produced by deep networks, encode semantic content primarily through the orientation of latent vectors rather than their norm, which can be affected by scale, activation statistics, or training dynamics. By normalizing all representations, we obtain a well-behaved similarity measure (cosine similarity) that directly reflects angular agreement between predicted and ground-truth codes, thereby providing a scale-invariant and geometry-preserving comparison.
Let $z \in \mathbb{R}^d$ denote a latent representation.
We map $z$ onto the unit hypersphere by $\ell_2$-normalization:

\begin{equation}
    \phi : \mathbb{R}^d \to \mathbb{S}^{d-1}, 
    \qquad
    \phi(z) = \frac{z}{\lVert z \rVert_2 + \varepsilon},
\end{equation}

where $\varepsilon > 0$ is a small constant added for numerical stability,
and $\mathbb{S}^{d-1} = \{ x \in \mathbb{R}^d : \lVert x \rVert_2 = 1 \}$ denotes 
the unit hypersphere in $\mathbb{R}^d$.

Given two latent vectors $z_{\text{render}}, z_{\text{target}} \in \mathbb{R}^d$, we first normalize them:
\begin{equation}
    \hat{r} = \phi\bigl(z_{\text{render}}\bigr)
    = \frac{z_{\text{render}}}{\lVert z_{\text{render}} \rVert_2 + \varepsilon},
    \qquad
    \hat{t} = \phi\bigl(z_{\text{target}}\bigr)
    = \frac{z_{\text{target}}}{\lVert z_{\text{target}} \rVert_2 + \varepsilon}.
\end{equation}
The similarity score used in our method is then the inner product between the two normalized vectors:
\begin{equation}
    \mathrm{sim}
    \;=\;
    \hat{r}^\top \hat{t}.
\end{equation}

Since $\lVert \hat{r} \rVert_2 = \lVert \hat{t} \rVert_2 = 1$ by construction, this inner product corresponds to the cosine of the angle $\theta$ between them:

\begin{equation}\label{eq:sim}
    \mathrm{sim}
    \;=\;
    \hat{r}^\top \hat{t}
    \;=\;
    \cos \theta,
    \qquad
    \theta = \angle(\hat{r}, \hat{t}).
\end{equation}

\paragraph{Pearson and Spearman correlation.}
Whereas cosine similarity collapses a latent vector to its direction, the Pearson and Spearman correlations ask whether the individual feature dimensions of the reconstruction and target latents agree. For two latents $z_{\text{render}}, z_{\text{target}} \in \mathbb{R}^d$, the Pearson coefficient is the linear correlation across the $d$ feature dimensions,
\begin{equation}
    \rho_{\text{P}} = \frac{\sum_k (z_{\text{render},k} - \bar{z}_{\text{render}})(z_{\text{target},k} - \bar{z}_{\text{target}})}{\sqrt{\sum_k (z_{\text{render},k} - \bar{z}_{\text{render}})^2}\,\sqrt{\sum_k (z_{\text{target},k} - \bar{z}_{\text{target}})^2}},
\end{equation}
and the Spearman coefficient $\rho_{\text{S}}$ is the same quantity computed on the ranks of the entries, capturing monotonic (possibly non-linear) agreement. Together they probe a stricter notion of equivalence than cosine: a reconstruction can align in aggregate direction (high cosine) yet have poorly coordinated feature dimensions (low $\rho_{\text{P}}$). The gap between $\rho_{\text{P}}$ and $\rho_{\text{S}}$ further indicates whether any mismatch is driven by a few outlier dimensions (large gap) or is a systematic non-linear warp (small gap). Note that these are distinct from the representational similarity analysis defined next, which correlates across views rather than across feature dimensions.

\paragraph{Representational Similarity Analysis (RSA)}
Complementarily, to the unit hypersphere similarity, RSA \cite{kriegeskorte2008representational} serves as a higher-level measure that evaluates not only pointwise similarity between individual latent codes but also the relational structure induced over an entire set of views. This global perspective is crucial when reconstructing a latent space: even if individual codes align in isolation, the reconstruction is only meaningful if the pairwise similarity structure, the geometry of the manifold, is preserved. It therefore provides a principled way to assess whether the reconstructed latent space captures the same representational organization as the ground-truth latent space, thereby complementing hyperspherical similarity.
Let $\{x_1,\dots,x_N\}$ denote a set of $N$ views (images). We consider two latent representations of these views, for example, produced by two different scenes:
\[
z_i^{(1)} = g_1(x_i) \in \mathbb{R}^{d_1},
\qquad
z_i^{(2)} = g_2(x_i) \in \mathbb{R}^{d_2},
\quad i = 1,\dots,N.
\]

We first map each latent vector onto the unit hypersphere (as in our similarity measure before) using $\ell_2$-normalization
\[
\hat{z}_i^{(m)} = 
\frac{z_i^{(m)}}{\lVert z_i^{(m)} \rVert_2 + \varepsilon},
\qquad
m \in \{1,2\}.
\]

For each representational space $m$, we then construct a  \emph{representational similarity matrix} (RSM) $S^{(m)} \in \mathbb{R}^{N \times N}$ with entries
\begin{equation}
    S^{(m)}_{ij}
    \;=\;
    k\!\bigl(\hat{z}_i^{(m)}, \hat{z}_j^{(m)}\bigr),
    \qquad
    k(a,b) = a^\top b,
    \qquad
    1 \le i,j \le N,
\end{equation}
i.e.\ the cosine similarity between the two normalized latent vectors. We define our dissimilarity matrix as 
$$D^{(m)}_{ij} = 1 - S^{(m)}_{ij}.$$

RSA quantifies how similar the \emph{geometry} of these two representational spaces is, by correlating their RSMs. Let $\operatorname{vec}_\triangle(\cdot)$ denote vectorization of the upper triangular (off-diagonal) entries of a matrix. The RSA score is
then defined as \begin{equation}
    \rho_{\mathrm{RSA}}
    \;=\;
    \mathrm{corr}\!\left(
        \operatorname{vec}_\triangle\bigl(D^{(1)}\bigr),
        \operatorname{vec}_\triangle\bigl(D^{(2)}\bigr)
    \right),
\end{equation}
where $\mathrm{corr}$ is a Pearson correlation coefficient.  A high value of $\rho_{\mathrm{RSA}}$ indicates that pairs of views that are similar (or dissimilar) in one latent space tend to be similarly related in the other latent space, i.e.\ the two spaces encode a similar representational structure over the set of views. 

\paragraph{Bayes factor}
For each criterion with a per-view decomposition, similarity, Pearson and Spearman correlation, we supplement the binary pass/fail decision with a rank-based Bayes factor BF${+0}$ \cite{rouder2009bayesian, morey2014simple} that quantifies the statistical evidence across views. Concretely, all $2n\text{views}$ similarity scores (reconstruction and baseline) are pooled and ranked, and BF${+0}$ is computed from the resulting two-sample $t$-statistic under a Jeffreys-Zellner-Siow prior (Cauchy scale $r = \sqrt{2}/2$), testing the directional hypothesis $H+$: mean rank of the reconstruction exceeds that of the baseline. Because each camera view constitutes an independent evaluation of the same scene, the per-view scores are approximately exchangeable under $H_0$, satisfying the assumptions of the rank-based test. RSA, by contrast, collapses across all views into a single scalar per epoch. The resulting epoch trajectory is strongly autocorrelated, and there is no semantically grounded pairing between reconstruction and baseline epochs. Treating training steps as independent observations would inflate the effective sample size and yield spuriously decisive evidence.

\paragraph{\normalfont\flip{}}
We evaluate image quality using \flip{} \cite{andersson2020flip}, a perceptual difference metric tailored for rendered images. \flip{} compares a rendered image against a reference ground truth and produces a per-pixel error map that approximates the difference a human observer would perceive when alternating between the two images. Unlike traditional pixel-based metrics such as Mean Squared Error or Peak Signal-to-Noise Ratio, which treat all pixel deviations as equally significant regardless of their visual impact, \flip{} weights errors according to human visual sensitivity, meaning that differences in perceptually salient regions contribute more to the score than numerically equivalent errors in regions where the eye is less discriminating. This distinction is important in rendering evaluation: a method may achieve a low mean squared error by spreading small errors uniformly across an image, yet produce visually jarring artifacts in high-contrast edges or fine detail, which \flip{} is specifically designed to penalize. 

An important consideration when using \flip{} in the context of path tracing is that it is a full-reference metric, it requires a clean, converged ground truth image to compare against. \flip{} is a full-reference image difference algorithm whose output is a new image indicating the magnitude of the perceived difference between two images at each pixel. This means \flip{} does not intrinsically distinguish between structured rendering errors (e.g., bias from an approximation) and stochastic noise arising from MC variance, both manifest as per-pixel deviations from the reference and are penalized similarly. In practice, this is handled by using a high-sample-count path-traced image as the reference ground truth, so that the Monte Carlo integration's variance in the reference is negligible. The test image, rendered with fewer samples or a different algorithm, is then compared against this converged reference, and the resulting \flip{} error map reflects both residual noise and any systematic bias. \flip{} does pay specific attention to point-like structures such as fireflies (bright noise artifacts), isolated pixels with colors that differ greatly from their surroundings, which commonly arise in path tracing due to high-variance Monte Carlo samples. However, \flip{}'s main limitation is its inability to detect masking, which can lead to overestimation of differences when they are not perceived because other image content masks them out. This is particularly relevant for Monte Carlo noise, since the human visual system is known to tolerate high-frequency stochastic noise more \cite{Georgiev2016, Chizhov2022, Korac2023} than structured artifacts of equal energy, a property that \flip{} only partially captures through its contrast sensitivity and perceptual color modeling, rather than through an explicit noise masking model.

\section{Results}\label{sec:results} 

\subsection{Material Reconstruction}\label{sec:results-material}
\begin{figure*}
    \centering
    \includegraphics[width=\linewidth]{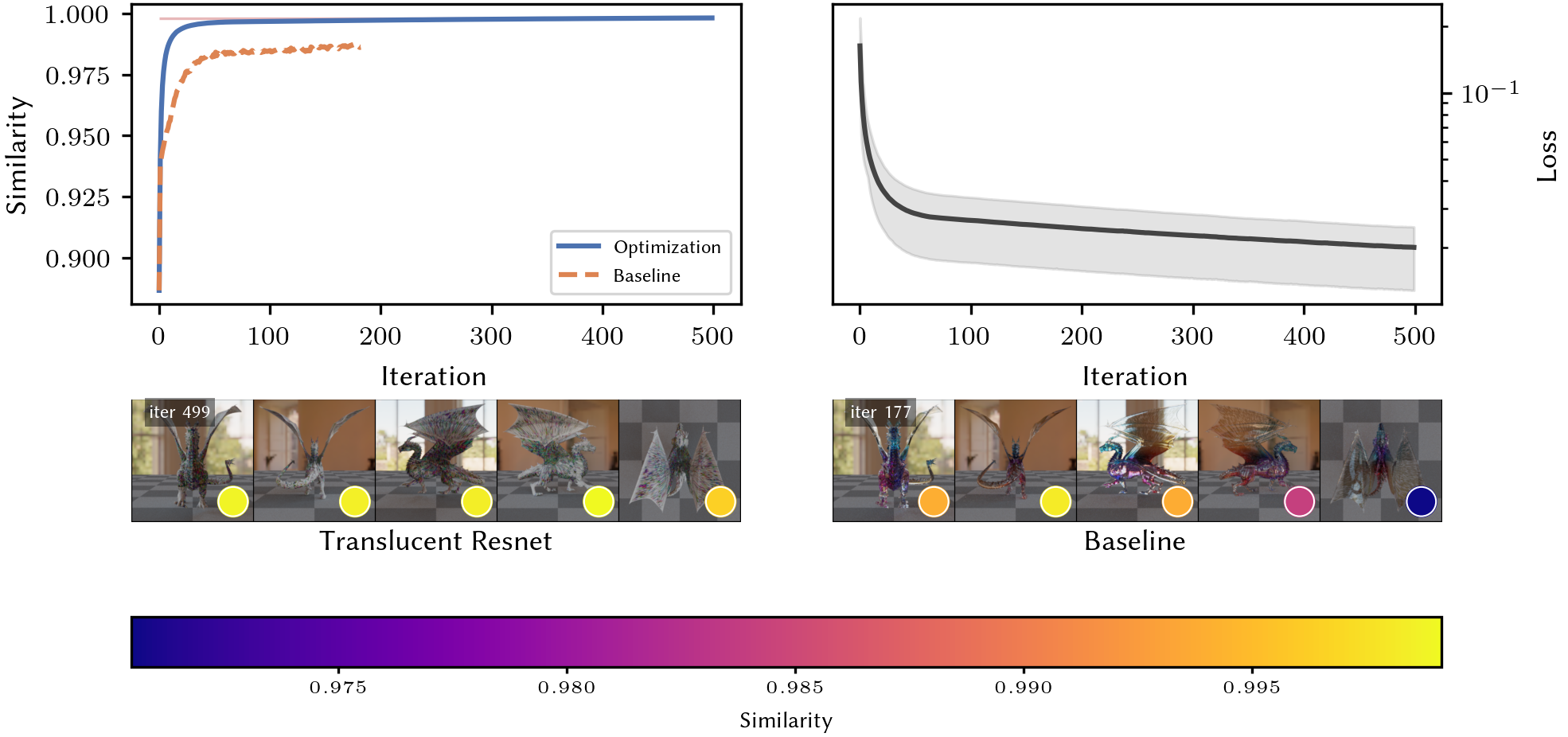}
    \caption{Translucent material reconstruction using ResNet's latent space. Left: cosine similarity to the target over optimization iterations --- the solid Optimization curve is the MRD reconstruction (network-driven), the dashed Baseline curve is the image-space reconstruction of the same scene; the reconstruction is a metamer where the solid curve reaches or exceeds the dashed one. Right: the corresponding loss curve (shading shows the min--max across views). The image strip below shows rendered views, and each colored dot encodes that view's similarity on the colorbar (brighter = higher). This is a case where the two outcomes dissociate: ResNet exceeds the baseline on every criterion (a metamer, primary outcome) yet ends up with a diffuse textured material that does not recover the ground-truth translucency (secondary outcome).}
    \label{fig:translucent-resnet}
\end{figure*}

\begin{figure*}
    \centering
    \includegraphics[width=\linewidth]{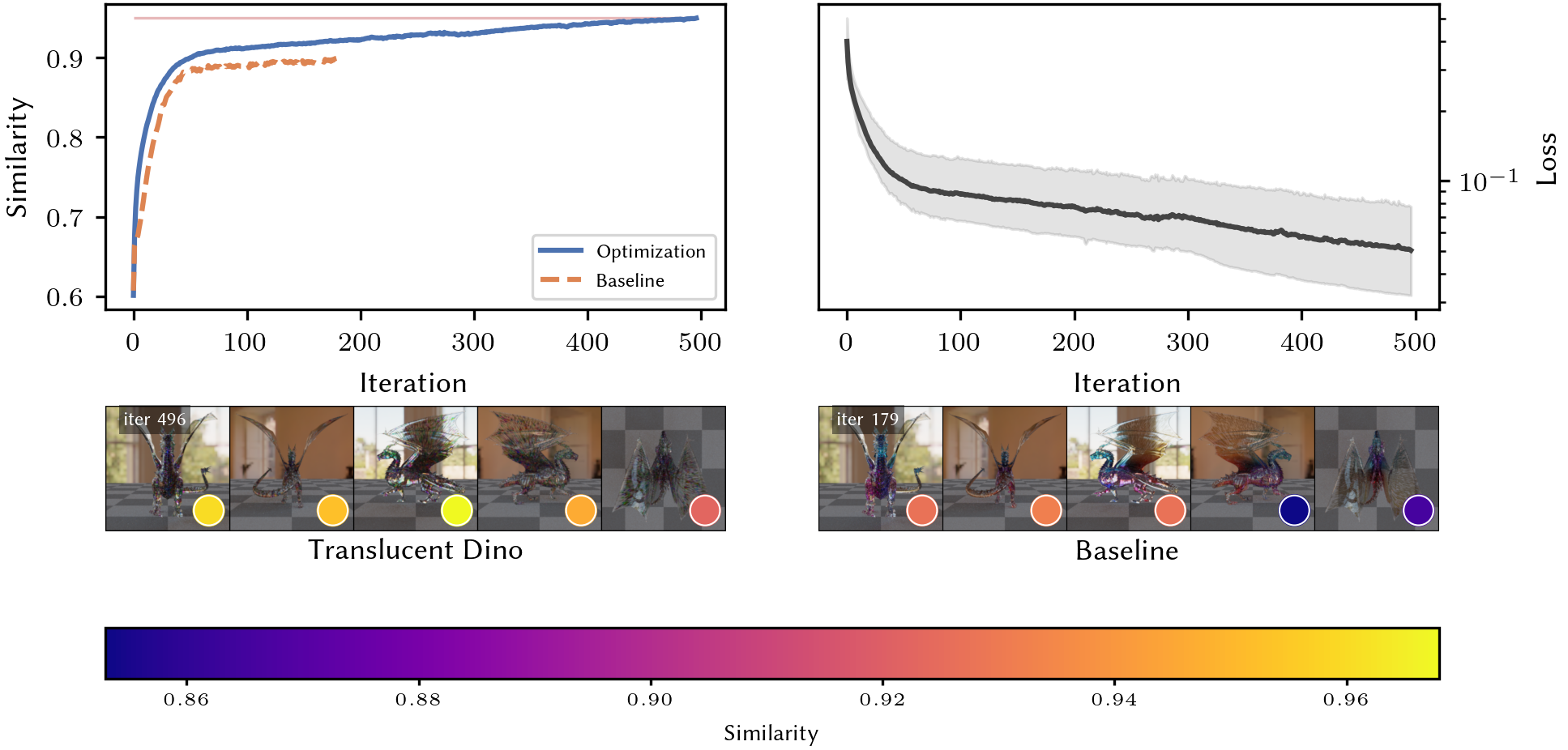}
    \caption{Translucent material reconstruction using DINO's latent space. Plot elements are as in Figure~\ref{fig:translucent-resnet} (solid Optimization = network-driven MRD, dashed Baseline = image-space reconstruction, loss curve with min--max shading, and per-view similarity encoded by the colored dots on the colorbar). DINO recovers parameters that afford translucency but still fails to reconstruct the correct texture --- again a metamer (primary) without full ground-truth recovery (secondary).}
    \label{fig:translucent-dino}
\end{figure*}

\begin{figure}
    \centering
    \includegraphics[width=\linewidth]{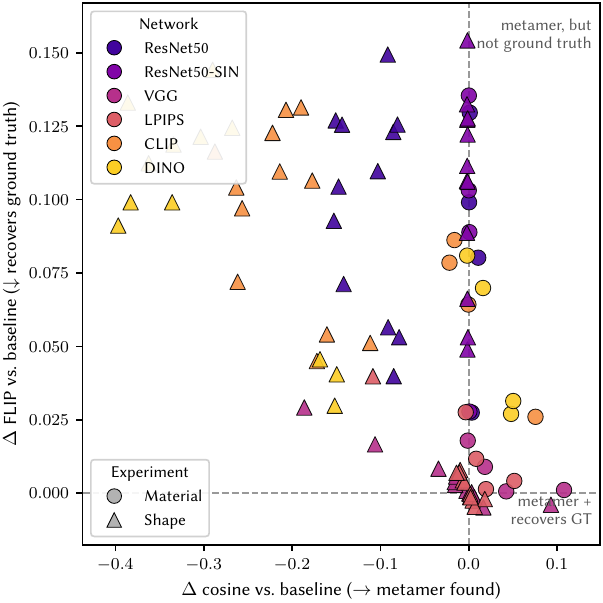}
    \caption{Separating the two outcomes. Each point is one experiment, placed by its metamer evidence ($x$: cosine $\Delta$ relative to baseline; points to the right of the dashed line are metamers) and its ground-truth recovery ($y$: \flip{} $\Delta$ relative to baseline; points below the dashed line have better-than-baseline image fidelity and so recover the ground truth more faithfully). Color encodes the network, marker shape distinguishes material (circles) from shape (triangles) experiments. The dissociation we emphasize in the text is the upper-right region: experiments that are metamers ($x > 0$) yet do not recover the ground truth ($y > 0$), populated mainly by material reconstructions (broad equivalence classes). Both coordinates are taken directly from Tables~\ref{tab:results_material} and~\ref{tab:results_shape_metamers}.}
    \label{fig:two-outcomes}
\end{figure}

\textbf{Primary outcome (metamers).} Across all material experiments, MRD reliably finds material metamers: 19 of 24 experiments meet at least one criterion, and 3 meet all four (Table~\ref{tab:results_material}). \textbf{Secondary outcome (ground-truth recovery).} However, these metamers frequently do not recover a \textit{Principled BSDF} parameter set that appears similar to the ground-truth material (as measured by \flip{}). The clearest examples are the translucent scenes: ResNet exceeds the baseline on every criterion yet converges to a diffuse, textured material with no translucency (Figure~\ref{fig:translucent-resnet}), and DINO recovers translucency but not the correct texture (Figure~\ref{fig:translucent-dino}). 
Therefore, MRD is effective at finding material metamers, and the networks' equivalence classes for material are broad rather than specific --- many physically different materials map to indistinguishable latents. Figure~\ref{fig:two-outcomes} plots this dissociation across all experiments: metamers (right of the dashed line) that nonetheless do not recover the ground truth. With that distinction in place, we report the per-metric pattern, summarized across all four metrics in Figure~\ref{fig:scatter}.

The \textit{LPIPS} perceptual metric was particularly effective for materials, consistently producing metamers across nearly all tested materials (Aurora, Brushed Metal, Diffuse, Translucent) with top similarity scores frequently exceeding $0.95$. Similarly, \textit{VGG} demonstrated strong performance for several materials. Crucially, cosine similarity, Pearson, and Spearman correlations agree closely across all material conditions: \textit{ResNet} achieves Pearson $\geq 0.998$ and Spearman $\geq 0.987$, while \textit{DINO} reaches Pearson $\geq 0.950$ and Spearman $\geq 0.938$, confirming that material optimization recovers not merely the direction of the latent vector but its full internal coordinate structure. RSA values further corroborate this picture: \textit{DINO} and \textit{ResNet} reach $\rho \geq 0.997$ across all four materials, indicating that the relational geometry across views is preserved to a near-ceiling degree. \textit{ResNet-SIN} is an exception to this pattern — despite achieving very high cosine similarity, its RSA is lower ($\rho = 0.42$–$0.86$), suggesting that its shape bias (from SIN training) disrupts the inter-view relational structure even when aggregate directional alignment is strong. Although a portion of the material experiments did not strictly meet the metamer criterion, many of these failures were marginal, with similarity gaps of $\Delta_{\text{sim}} < 0.02$ for several \textit{ResNet-SIN} conditions, indicating that small optimization noise rather than representational mismatch prevented formal qualification as metamers. Overall, material reconstruction is robust across networks at all levels of latent structure — per-dimension (Pearson, Spearman), directional (cosine), and relational (RSA) — with LPIPS, VGG, DINO, and ResNet providing the highest fidelity.


\begin{table*}[htbp]
    \caption{Material reconstruction results sorted by metamer count (\#). Each metric group shows: top score, signed difference from baseline ($\Delta$; Cosine = top$-$baseline, RSA = RSA$-$baseline RSA), and Bayes factor ($\log_{10}$) or significance. \flip{} shows the best (minimum) perceptual error and its signed difference from the baseline \flip{} ($\Delta < 0$ = better image quality than baseline). \colorbox{green!20}{Green} indicates the criterion was met. ``---'' denotes metrics unavailable for LPIPS/VGG. RSA scores are significant (s) if \( p < 0.05 \) else not significant (n.s.).}
    \label{tab:results_material}
    \setlength{\tabcolsep}{3pt}
    \begin{tabular}{lc rrr rrr rr rr rr}
        \toprule
        \multirow{2}{*}{Experiment} & \multirow{2}{*}{\#}
                                    & \multicolumn{3}{c}{Cosine Similarity}
                                    & \multicolumn{3}{c}{RSA}
                                    & \multicolumn{2}{c}{Pearson}
                                    & \multicolumn{2}{c}{Spearman}
                                    & \multicolumn{2}{c}{\flip{}} \\
                                    \cmidrule(lr){3-5}\cmidrule(lr){6-8}\cmidrule(lr){9-10}\cmidrule(lr){11-12}\cmidrule(lr){13-14}
                                    & & Score & $\Delta$ & BF & Score & $\Delta$ & Sig & Score & BF & Score & BF & Score & $\Delta$ \\
                                    \midrule
        Brushed Metal ResNet50   & 4 & \cellcolor{green!20}0.9998 & $+0.004$ & $-0.06$ & \cellcolor{green!20}1.000 & $+0.011$ & s & \cellcolor{green!20}0.9997  & $1.72$ & \cellcolor{green!20}0.9974  & $0.46$ & 0.0249 & $+0.005$ \\
        Diffuse ResNet50         & 4 & \cellcolor{green!20}0.9997 & $+0.000$ & $-0.06$ & \cellcolor{green!20}1.000  & $+0.059$ & s & \cellcolor{green!20}0.9995  & $0.21$ & \cellcolor{green!20}0.9969  & $-0.14$ & 0.0628 & $+0.053$ \\
        Translucent ResNet50     & 4 & \cellcolor{green!20}0.9983 & $+0.011$ & $1.20$ & \cellcolor{green!20}0.999  & $+0.068$ & s & \cellcolor{green!20}0.9975  & $1.72$ & \cellcolor{green!20}0.9882  & $1.20$ & 0.0804 & $+0.049$ \\
        \midrule
        Aurora ResNet50          & 3 & \cellcolor{green!20}0.9988 & $+0.001$ & $-0.06$ & \cellcolor{green!20}0.997 & $+0.034$ & s & \cellcolor{green!20}0.9982  & $0.10$ & 0.9898  & $-0.47$ & 0.0889 & $+0.071$ \\
        Translucent ResNet50-SIN & 3 & \cellcolor{green!20}0.9991 & $+0.001$ & $1.72$ & 0.807 & $-0.103$ & s & \cellcolor{green!20}0.9718 & $-0.47$ & \cellcolor{green!20}0.9561  & $-0.54$ & 0.0847 & $+0.054$ \\
        \midrule
        Aurora DINO              & 2 & \cellcolor{green!20}0.9851 & $+0.016$ & $0.46$ & \cellcolor{green!20}0.999 & $+0.024$ & s & 0.9851 & $-1.21$ & 0.9805 & $-1.21$ & 0.0418 & $+0.024$ \\
        Brushed Metal CLIP       & 2 & \cellcolor{green!20}0.9514 & $+0.076$ & $0.64$ & \cellcolor{green!20}0.953 &  $+0.010$ & s & 0.9513 & $-1.37$ & 0.9215  & $-1.37$ & 0.0245 & $+0.005$ \\
        Brushed Metal DINO       & 2 & \cellcolor{green!20}0.9878 & $+0.048$ & $1.20$ & \cellcolor{green!20}0.999 &  $+0.021$ & s & 0.9877  & $-0.62$ & 0.9849  & $-0.34$ & 0.0256 & $+0.006$ \\
        Diffuse ResNet50-SIN     & 2 & \cellcolor{green!20}0.9994 & $+0.000$ & $-0.14$ & \cellcolor{green!20}0.858 &  $+0.264$ & s & 0.9804  & $-1.56$ & 0.9703  & $-1.56$ & 0.0661 & $+0.056$ \\
        Translucent DINO         & 2 & \cellcolor{green!20}0.9497 & $+0.050$ & $0.87$ & \cellcolor{green!20}0.996 & $+0.010$ & s & 0.9496  & $-0.54$ & 0.9384  & $-0.54$ & 0.0432 & $+0.012$ \\
        \midrule
        Aurora LPIPS             & 1 & \cellcolor{green!20}0.9948 & $+0.008$ & $0.64$ & \multicolumn{3}{c}{---} & \multicolumn{2}{c}{---} & \multicolumn{2}{c}{---} & \cellcolor{green!20}0.0168 & $-0.001$ \\
        Aurora ResNet50-SIN      & 1 & \cellcolor{green!20}0.9993 & $+0.000$ & $0.21$ & 0.458 & $-0.382$ & n.s. & 0.9767 &  $-1.56$ & 0.9630 & $-1.56$ & 0.0919 & $+0.074$ \\
        Aurora VGG               & 1 & \cellcolor{green!20}0.9890 & $+0.018$ & $0.64$ & \multicolumn{3}{c}{---} & \multicolumn{2}{c}{---} & \multicolumn{2}{c}{---} & \cellcolor{green!20}0.0163 & $-0.001$ \\
        Brushed Metal LPIPS      & 1 & \cellcolor{green!20}0.9941 & $+0.020$ & $0.46$ & \multicolumn{3}{c}{---} & \multicolumn{2}{c}{---} & \multicolumn{2}{c}{---} & \cellcolor{green!20}0.0167 & $-0.003$ \\
        Brushed Metal ResNet50-SIN & 1 & \cellcolor{green!20}0.9997  & $+0.001$ & $1.72$ & 0.638 &  $-0.173$ & s & 0.9887  & $-0.62$ & 0.9821  & $-0.62$ & 0.0264 & $+0.007$ \\
        Brushed Metal VGG        & 1 & \cellcolor{green!20}0.9865  & $+0.042$ & $0.46$ & \multicolumn{3}{c}{---} & \multicolumn{2}{c}{---} & \multicolumn{2}{c}{---} & \cellcolor{green!20}0.0166 & $-0.003$ \\
        Diffuse DINO             & 1 & 0.9857  & $-0.002$ & $-0.54$ & \cellcolor{green!20}0.997  & $+0.013$ & s & 0.9857  & $-1.56$ & 0.9829  & $-1.37$ & 0.0526 & $+0.042$ \\
        Translucent LPIPS        & 1 & \cellcolor{green!20}0.9492  & $+0.052$ & $0.32$ & \multicolumn{3}{c}{---} & \multicolumn{2}{c}{---} & \multicolumn{2}{c}{---} & \cellcolor{green!20}0.0308 & $-0.000$ \\
        Translucent VGG          & 1 & \cellcolor{green!20}0.8904  & $+0.108$ & $0.32$ & \multicolumn{3}{c}{---} & \multicolumn{2}{c}{---} & \multicolumn{2}{c}{---} & \cellcolor{green!20}0.0299 & $-0.001$ \\
        \midrule
        Aurora CLIP              & 0 & 0.9066  & $-0.016$ & $-0.34$ & 0.915 & $-0.073$ & s & 0.9064  & $-1.56$ & 0.8480  & $-1.56$ & 0.0537 & $+0.036$ \\
        Diffuse CLIP             & 0 & 0.9324  & $-0.022$ & $-0.78$ & 0.916  & $-0.066$ & s & 0.9322  & $-1.56$ & 0.8948  & $-1.56$ & 0.0504 & $+0.040$ \\
        Diffuse LPIPS            & 0 & 0.9914  & $-0.003$ & $-0.69$ & \multicolumn{3}{c}{---} & \multicolumn{2}{c}{---} & \multicolumn{2}{c}{---} & 0.0224 & $+0.012$ \\
        Diffuse VGG              & 0 & 0.9874  & $-0.001$ & $-0.47$ & \multicolumn{3}{c}{---} & \multicolumn{2}{c}{---} & \multicolumn{2}{c}{---} & 0.0176 & $+0.007$ \\
        Translucent CLIP         & 0 & 0.7926  & $-0.000$ & $-0.41$ & 0.770  & $-0.205$ & s & 0.7920  & $-1.56$ & 0.6620  & $-1.56$ & 0.0683 & $+0.037$ \\
        \bottomrule
    \end{tabular}
\end{table*}
Across our different scenes, we were able to find metamers for most of the networks tested (Table~\ref{tab:results_material}).
The translucent reconstructions, ResNet and CLIP, had examples that did not qualify as metamers under our definition. 

\paragraph{Material-specific Effects and Difficulty Regimes}
Different material classes show systematically different recoverability. Metallic and specular materials (brushed-metal, aurora) tend to produce strong global image cues (high-contrast highlights, anisotropic streaks) that generate coherent gradients and are thus easier to recover to baseline similarity. Diffuse and translucent materials can be more challenging: they sometimes produce broader similarity gaps or lower RSA values, likely because their appearance depends on subtle indirect-lighting interactions or transmission effects that are harder to disambiguate from other scene factors. Where translucency interacts with geometry and indirect illumination, the optimization needs a larger number of samples and iterations to reach the same representational fidelity.

\paragraph{Noise, Clipping and other Practical Considerations}
Material reconstruction relies on unbiased path tracing and, therefore, is subject to Monte Carlo noise in both forward rendering and gradient estimates. We found that (1) parameter clipping to the physically valid domain is essential to avoid unstable and physically implausible updates, and (2) increasing sample count can reduce noisy gradients and improve final alignment. (3) For complex scenes, truncating the maximum number of paths per ray can also affect final results, especially for translucent material, where light can refract and scatter multiple times within the same object. (4) Non-translucent materials and texture reconstructions require complete view coverage to be reconstructed properly. The spatial resolution can be further reduced to allow for more ray intersections.

\subsection{Shape Reconstruction}\label{sec:results-shape}

\begin{figure*}
    \centering
    \includegraphics[width=\linewidth]{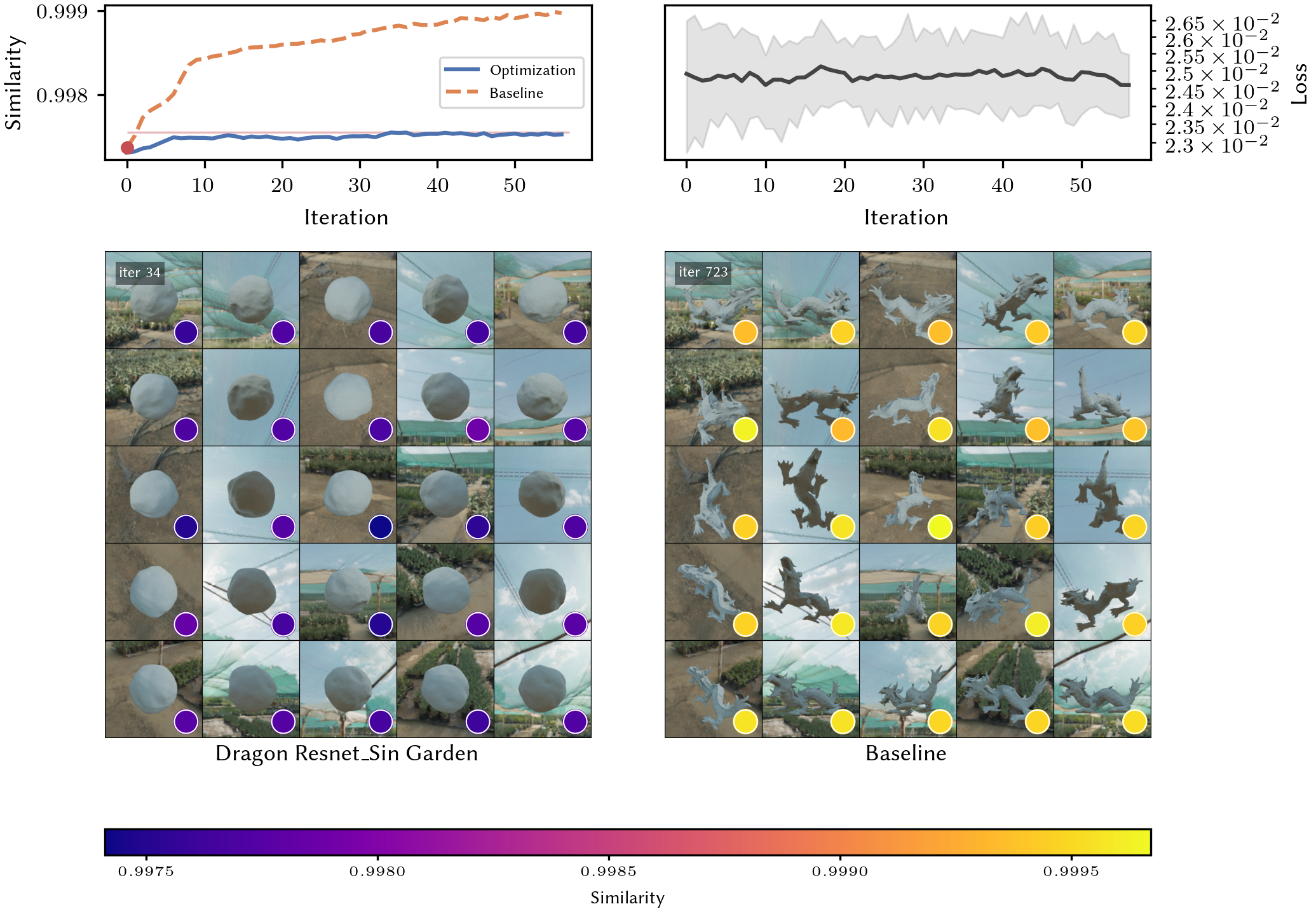}
    \caption{Shape reconstruction using ResNet-SIN's latent space. This figure should be read as a cautionary case, not a success: the cosine similarity (solid Optimization curve) is high and almost flat from the very first iteration, and the loss barely moves --- not because the optimization solved the problem, but because ResNet-SIN maps almost any object to a near-identical direction in latent space, so cosine is already saturated before optimization begins (the baseline is correspondingly high too). The reconstruction therefore passes the cosine criterion essentially trivially while the structural metrics (Pearson, Spearman, RSA; Table~\ref{tab:results_shape_metamers}) remain low and the rendered shape does not match the target. In other words, ResNet-SIN's bias enlarges the equivalence class so much that cosine metamerism becomes uninformative --- which is why we rely on the stricter structural metrics.}
    \label{fig:resnetsin-shape}
\end{figure*}

\begin{figure*}
    \centering
    \includegraphics[width=\linewidth]{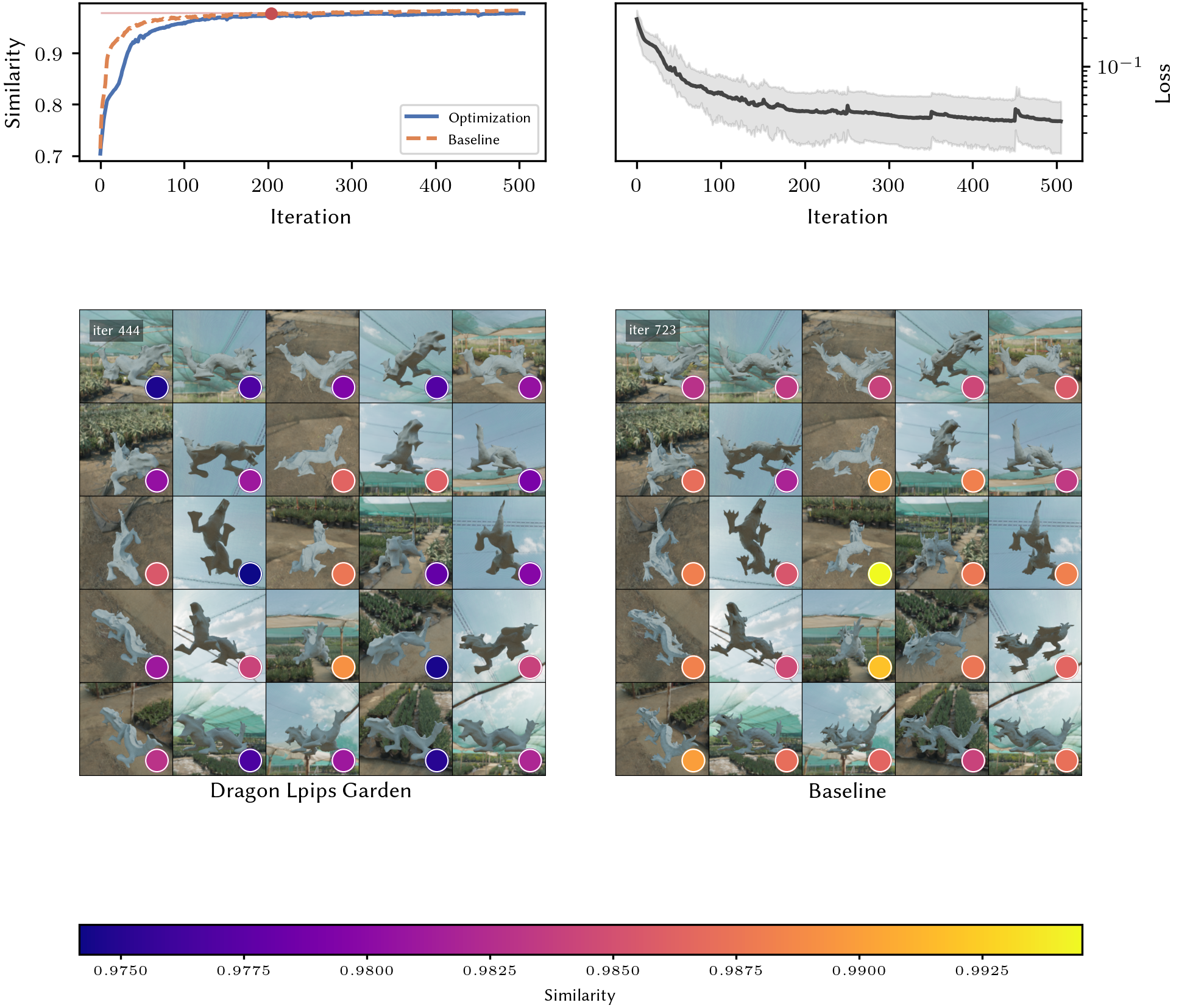}
    \caption{Shape reconstruction of the Dragon in the Garden environment using LPIPS as the optimization objective; LPIPS is our strongest objective for shape. Top left: similarity to the target over optimization iterations --- the solid Optimization curve is the MRD reconstruction (driven by LPIPS), the dashed Baseline curve is the image-space reconstruction of the same scene, and the red dot marks the peak epoch we report. Top right: the corresponding loss curve (shading shows the min--max across views). Bottom: the same set of camera views for the MRD reconstruction (left, at its peak iteration) and the baseline (right); each colored dot encodes that view's similarity on the colorbar (brighter = higher; note the narrow $0.975$--$0.993$ range, so all views are already highly similar). Both reconstruction and baseline reach very high similarity, and the MRD reconstruction tracks the near-ideal baseline closely while only marginally falling short of formally exceeding it (Table~\ref{tab:results_shape_metamers}). Crucially, unlike the trivially-high but uninformative similarities of the ResNet-SIN case (Figure~\ref{fig:resnetsin-shape}), here the high scores coincide with genuine recovery: the rendered dragon visually matches the target across viewpoints.}
    \label{fig:lpips-shape}
\end{figure*}

As with material, we separate the two outcomes. \textbf{Primary outcome (metamers).} Shape is the harder case, and here the primary outcome usually fails: only 33 of 72 shape experiments meet even one criterion, and far fewer meet more than one (Table~\ref{tab:results_shape_metamers}), so more often than not a shape reconstruction is not a metamer, and the additional metrics expose failures that cosine similarity alone conceals. \textbf{Secondary outcome (ground-truth recovery).} Here shape differs instructively from material: rather than dissociating, the two outcomes are largely aligned. In the cases where a reconstruction does reach the baseline --- the perceptual losses LPIPS and VGG, and the geometrically simple Suzanne --- it also tends to recover the target geometry, with the rendered object visually matching the target across views (e.g.\ the dragon of Figure~\ref{fig:lpips-shape}). The telling exception is ResNet-SIN, whose near-perfect cosine similarity is trivially high yet does not recover the true geometry  (Figure~\ref{fig:resnetsin-shape}). 
We therefore read the successes, concentrated in the perceptual losses and simple geometry, as genuine proof-of-concept cases and highlight them below. 

\textit{LPIPS} remained the strongest single metric for shapes, achieving metamers across several objects (Dog, Dragon, Lion Statue, Suzanne), and \textit{VGG} also succeeded in several cases, reflecting its sensitivity to mid-level image structure. However, \textit{ResNet} reveals a pronounced dissociation between directional and structural fidelity: cosine similarity reaches $0.83$–$0.91$ on Dog and Dragon scenes, yet Pearson correlation drops to $0.44$–$0.76$ and Spearman to $0.34$–$0.67$, indicating that individual feature dimensions are poorly coordinated even when the aggregate latent direction is roughly aligned. \textit{ResNet-SIN} displays the most extreme version of this pattern (Figure~\ref{fig:resnetsin-shape}) — its cosine similarity is near-perfect ($\approx 0.997$) across all shape experiments, yet Pearson plateaus at $\approx 0.92$ and Spearman at $\approx 0.88$, a ceiling imposed by its shape bias, which aligns global feature directions while leaving the internal latent structure underspecified. RSA tells a similarly degraded story: values for Dog, Dragon, and Lion Statue fall in the $0.28$–$0.66$ range and, in the majority of conditions, remain \emph{below} the image-space mean absolute error baseline RSA, meaning a simple pixel reconstruction preserves inter-view relational geometry better than the perceptual-loss optimization. Suzanne is a consistent exception: its simpler convex geometry allows the perceptual losses to recover relational structure that exceeds the baseline (e.g., \textit{ResNet-SIN Hallstatt}: RSA $= 0.851$ vs.\ baseline $0.598$), pointing to geometric complexity as the primary limiting factor rather than the choice of perceptual objective. The narrow Pearson--Spearman gap across all shape conditions ($0.05$--$0.10$) suggests that the latent distributions are monotonically but non-linearly related, with no strong outlier distortion — the mismatch is systematic rather than view-specific. 

Taken together, these results indicate that shape reconstruction is more susceptible to optimization noise and the high dimensionality of the geometry space, and that perceptual losses optimize individual-view fidelity at the expense of the global representational geometry, a failure mode invisible to per-view metrics alone.

\begin{table*}[htbp]
    \caption{Shape reconstruction results sorted by metamer count (\#). Each metric group shows: top score, signed difference from baseline ($\Delta$; Cosine = top$-$baseline sim, RSA = RSA$-$baseline RSA), and Bayes factor (BF$\log_{10}$) or significance. \flip{} shows the best (minimum) perceptual error and its signed difference from the baseline \flip{} ($\Delta < 0$ = better image quality than baseline). \colorbox{green!20}{Green} shading indicates the criterion was met. ``---'' denotes metrics unavailable for LPIPS/VGG. For RSA all values are significant (\( p < 0.05 \)).}
    \label{tab:results_shape_metamers}
    \centering
    \small
    \setlength{\tabcolsep}{2pt}
    \renewcommand{\arraystretch}{0.85}
    \begin{tabular}{lc rrr rr rr rr rr}
        \toprule
        \multirow{2}{*}{Experiment} & \multirow{2}{*}{\#}
                                    & \multicolumn{3}{c}{Cosine Similarity}
                                    & \multicolumn{2}{c}{RSA}
                                    & \multicolumn{2}{c}{Pearson}
                                    & \multicolumn{2}{c}{Spearman}
                                    & \multicolumn{2}{c}{\flip{}} \\
                                    \cmidrule(lr){3-5}\cmidrule(lr){6-7}\cmidrule(lr){8-9}\cmidrule(lr){10-11}\cmidrule(lr){12-13}
                                    & & Score & $\Delta$ & BF & Score & $\Delta$  & Score & BF & Score & BF & Score & $\Delta$ \\
                                    \midrule
        Dog CLIP Garden          & 1 & 0.6957  & $-0.257$ & $-2.44$ & \cellcolor{green!20}0.404  & $+0.026$  & 0.6927 &  $-2.44$ & 0.5680 &  $-2.44$ & 0.0787 & $+0.070$ \\
        Dog CLIP Hallstatt       & 1 & 0.7024  & $-0.262$ & $-2.44$ & \cellcolor{green!20}0.545  & $+0.260$  & 0.6975 &  $-2.44$ & 0.5746 &  $-2.44$ & 0.0395 & $+0.034$ \\
        Dog DINO Garden          & 1 & 0.5990  & $-0.383$ & $-2.44$ & \cellcolor{green!20}0.525  & $+0.022$  & 0.5445 &  $-2.44$ & 0.4777 &  $-2.44$ & 0.0866 & $+0.078$ \\
        Dog LPIPS Garden         & 1 & \cellcolor{green!20}0.9895  & $+0.004$ & $11.15$ & \multicolumn{2}{c}{---} & \multicolumn{2}{c}{---} & \multicolumn{2}{c}{---} & \cellcolor{green!20}0.0073 & $-0.002$ \\
        Dog LPIPS Hallstatt      & 1 & \cellcolor{green!20}0.9919  & $+0.002$ & $3.86$ & \multicolumn{2}{c}{---} & \multicolumn{2}{c}{---} & \multicolumn{2}{c}{---} & \cellcolor{green!20}0.0051 & $-0.000$ \\
        Dog LPIPS Skybox         & 1 & \cellcolor{green!20}0.9685  & $+0.018$ & $8.38$ & \multicolumn{2}{c}{---} & \multicolumn{2}{c}{---} & \multicolumn{2}{c}{---} & 0.0065 & $+0.001$ \\
        Dog ResNet50-SIN Garden  & 1 & 0.9974  & $-0.002$ & $-2.44$ & \cellcolor{green!20}0.366  & $+0.162$ &  0.9169  & $-2.44$ & 0.8789   & $-2.44$ & 0.0859 & $+0.077$ \\
        Dog ResNet50-SIN Hallstatt & 1 & 0.9975  & $-0.002$ & $-2.44$ & \cellcolor{green!20}0.456  & $+0.008$ & 0.9209  & $-2.44$ & 0.8823  & $-2.44$ & 0.0542 & $+0.049$ \\
        Dog VGG Skybox           & 1 & \cellcolor{green!20}0.9711 & $+0.093$ & $13.20$ & \multicolumn{2}{c}{---} & \multicolumn{2}{c}{---} & \multicolumn{2}{c}{---} & 0.0054 & $+0.000$ \\
        Dragon CLIP Garden       & 1 & 0.6460 & $-0.287$ & $-2.44$ & \cellcolor{green!20}0.658  & $+0.352$  & 0.6306  & $-2.44$ & 0.4937  & $-2.44$ & 0.0673 & $+0.056$ \\
        Dragon CLIP Hallstatt    & 1 & 0.6398 & $-0.263$ & $-2.44$ & \cellcolor{green!20}0.369  & $+0.096$  & 0.6243  & $-2.44$ & 0.4807  & $-2.44$ & 0.0578 & $+0.047$ \\
        Dragon ResNet50 Skybox   & 1 & 0.8368 & $-0.143$ & $-2.44$ & \cellcolor{green!20}0.525  & $+0.111$  & 0.6787  & $-2.44$ & 0.5896  & $-2.44$ & 0.0885 & $+0.074$ \\
        Dragon ResNet50-SIN Garden & 1 & 0.9975   & $-0.002$ & $-2.44$ & \cellcolor{green!20}0.346  & $+0.023$  & 0.9195 & $-2.44$ & 0.8812  & $-2.44$ & 0.0924 & $+0.081$ \\
        Dragon ResNet50-SIN Hallstatt & 1 & 0.9975   & $-0.002$ & $-2.44$ & \cellcolor{green!20}0.440   & $+0.007$  & 0.9212   & $-2.44$ & 0.8836   & $-2.44$ & 0.0657 & $+0.055$ \\
        Lion Statue CLIP Garden  & 1 & 0.6563  & $-0.207$ & $-2.40$ & \cellcolor{green!20}0.613 & $+0.043$  & 0.6092 & $-2.44$ & 0.4597  & $-2.44$ & 0.0897 & $+0.073$ \\
        Lion Statue VGG Hallstatt & 1 & \cellcolor{green!20}0.9124 & $+0.001$ & $-0.27$ & \multicolumn{2}{c}{---} & \multicolumn{2}{c}{---} & \multicolumn{2}{c}{---} & \cellcolor{green!20}0.0211 & $-0.005$ \\
        Suzanne CLIP Garden      & 1 & 0.7773  & $-0.161$ & $-1.86$ & \cellcolor{green!20}0.586  & $+0.370$  & 0.7488  & $-1.86$ & 0.6433  & $-1.86$ & 0.0568 & $+0.042$ \\
        Suzanne CLIP Hallstatt   & 1 & 0.7707  & $-0.172$ & $-1.86$ & \cellcolor{green!20}0.495  & $+0.261$  & 0.7477  & $-1.86$ & 0.6409  & $-1.86$ & 0.0352 & $+0.026$ \\
        Suzanne CLIP Skybox      & 1 & 0.7486  & $-0.111$ & $-0.72$ & \cellcolor{green!20}0.602  & $+0.213$  & 0.7262  & $-1.86$ & 0.6196  & $-1.86$ & 0.0488 & $+0.034$ \\
        Suzanne DINO Garden      & 1 & 0.8245  & $-0.149$ & $-1.86$ & \cellcolor{green!20}0.603  & $+0.382$  & 0.7644  & $-1.86$ & 0.7182  & $-1.86$ & 0.0465 & $+0.032$ \\
        Suzanne DINO Hallstatt   & 1 & 0.8262  & $-0.152$ & $-1.86$ & \cellcolor{green!20}0.558  & $+0.309$  & 0.7628  & $-1.86$ & 0.7170  & $-1.86$ & 0.0279 & $+0.019$ \\
        Suzanne DINO Skybox      & 1 & 0.7236  & $-0.168$ & $-1.04$ & \cellcolor{green!20}0.596  & $+0.276$  & 0.6701  & $-1.86$ & 0.6114  & $-1.86$ & 0.0401 & $+0.025$ \\
        Suzanne LPIPS Garden     & 1 & \cellcolor{green!20}0.9904  & $+0.007$ & $2.49$ & \multicolumn{2}{c}{---} & \multicolumn{2}{c}{---} & \multicolumn{2}{c}{---} & \cellcolor{green!20}0.0118 & $-0.003$ \\
        Suzanne LPIPS Hallstatt  & 1 & \cellcolor{green!20}0.9896  & $+0.002$ & $0.68$ & \multicolumn{2}{c}{---} & \multicolumn{2}{c}{---} & \multicolumn{2}{c}{---} & 0.0095 & $+0.001$ \\
        Suzanne LPIPS Skybox     & 1 & \cellcolor{green!20}0.9563  & $+0.002$ & $0.78$ & \multicolumn{2}{c}{---} & \multicolumn{2}{c}{---} & \multicolumn{2}{c}{---} & \cellcolor{green!20}0.0129 & $-0.002$ \\
        Suzanne ResNet50 Garden  & 1 & 0.9015  & $-0.091$ & $-1.86$ & \cellcolor{green!20}0.558   & $+0.076$  & 0.6433  & $-1.86$ & 0.5188   & $-1.86$ & 0.0539 & $+0.040$ \\
        Suzanne ResNet50 Hallstatt & 1 & 0.9090  & $-0.085$ & $-1.86$ & \cellcolor{green!20}0.539   & $+0.237$  & 0.6676  & $-1.86$ & 0.5591  & $-1.86$ & 0.0340 & $+0.025$ \\
        Suzanne ResNet50-SIN Garden & 1 & 0.9978   & $-0.001$ & $-1.86$ & \cellcolor{green!20}0.662   & $+0.111$  & 0.9273   & $-1.86$ & 0.8917   & $-1.86$ & 0.0632 & $+0.049$ \\
        Suzanne ResNet50-SIN Hallstatt & 1 & 0.9979  & $-0.001$ & $-1.86$ & \cellcolor{green!20}0.851   & $+0.253$  & 0.9318   & $-1.86$ & 0.8978   & $-1.86$ & 0.0404 & $+0.031$ \\
        Suzanne ResNet50-SIN Skybox & 1 & 0.9978    & $-0.001$ & $-1.04$ & \cellcolor{green!20}0.613    & $+0.302$  & 0.9286   & $0.23$ & 0.8932   & $-0.95$ & 0.0488 & $+0.034$ \\
        Suzanne VGG Garden       & 1 & \cellcolor{green!20}0.9742  & $+0.017$ & $2.49$ & \multicolumn{2}{c}{---} & \multicolumn{2}{c}{---} & \multicolumn{2}{c}{---} & \cellcolor{green!20}0.0117 & $-0.003$ \\
        Suzanne VGG Hallstatt    & 1 & \cellcolor{green!20}0.9718  & $+0.003$ & $0.29$ & \multicolumn{2}{c}{---} & \multicolumn{2}{c}{---} & \multicolumn{2}{c}{---} & 0.0098 & $+0.001$ \\
        Suzanne VGG Skybox       & 1 & \cellcolor{green!20}0.8924  & $+0.006$ & $0.78$ & \multicolumn{2}{c}{---} & \multicolumn{2}{c}{---} & \multicolumn{2}{c}{---} & \cellcolor{green!20}0.0134 & $-0.001$ \\
        \midrule
        Dog CLIP Skybox          & 0 & 0.6766 & $-0.177$ & $-2.44$ & 0.461  & $-0.125$  & 0.6662  & $-2.44$ & 0.5376  & $-2.44$ & 0.0669 & $+0.062$ \\
        Dog DINO Hallstatt       & 0 & 0.5908 & $-0.397$ & $-2.44$ & 0.389  & $-0.119$  & 0.5384  & $-2.44$ & 0.4718  & $-2.44$ & 0.0565 & $+0.051$ \\
        Dog DINO Skybox          & 0 & 0.5739 & $-0.336$ & $-2.44$ & 0.276  & $-0.343$  & 0.5395  & $-2.44$ & 0.4723  & $-2.44$ & 0.0677 & $+0.062$ \\
        Dog ResNet50 Garden      & 0 & 0.8345 & $-0.153$ & $-2.44$ & 0.379  & $-0.282$  & 0.4660  & $-2.44$ & 0.3734  & $-2.44$ & 0.0792 & $+0.070$ \\
        Dog ResNet50 Hallstatt   & 0 & 0.8493 & $-0.142$ & $-2.44$ & 0.462  & $-0.214$  & 0.4460  & $-2.44$ & 0.3395  & $-2.44$ & 0.0493 & $+0.044$ \\
        Dog ResNet50 Skybox      & 0 & 0.8004 & $-0.147$ & $-2.44$ & 0.345  & $-0.144$  & 0.4651  & $-2.44$ & 0.3667  & $-2.44$ & 0.0669 & $+0.062$ \\
        Dog ResNet50-SIN Skybox  & 0 & 0.9974 & $-0.001$ & $-2.43$ & 0.304  & $-0.261$  & 0.9152  & $-1.96$ & 0.8752  & $-2.44$ & 0.0680 & $+0.063$ \\
        Dog VGG Garden           & 0 & 0.9598 & $-0.002$ & $3.62$ & \multicolumn{2}{c}{---} & \multicolumn{2}{c}{---} & \multicolumn{2}{c}{---} & 0.0100 & $+0.001$ \\
        Dog VGG Hallstatt        & 0 & 0.9601 & $-0.015$ & $-1.50$ & \multicolumn{2}{c}{---} & \multicolumn{2}{c}{---} & \multicolumn{2}{c}{---} & 0.0073 & $+0.002$ \\
        Dragon CLIP Skybox       & 0 & 0.6362 & $-0.222$ & $-2.44$ & 0.410  & $-0.212$  & 0.6154  & $-2.44$ & 0.4733  & $-2.44$ & 0.0881 & $+0.074$ \\
        Dragon DINO Garden       & 0 & 0.5882 & $-0.386$ & $-2.44$ & 0.492  & $-0.018$  & 0.4965  & $-2.44$ & 0.4251  & $-2.44$ & 0.0935 & $+0.082$ \\
        Dragon DINO Hallstatt    & 0 & 0.5989 & $-0.362$ & $-2.44$ & 0.418  & $-0.163$  & 0.5139  & $-2.44$ & 0.4465  & $-2.44$ & 0.0684 & $+0.058$ \\
        Dragon DINO Skybox       & 0 & 0.5886 & $-0.334$ & $-2.44$ & 0.444  & $-0.182$  & 0.5201  & $-2.44$ & 0.4510  & $-2.44$ & 0.0920 & $+0.078$ \\
        Dragon LPIPS Garden      & 0 & 0.9779 & $-0.007$ & $-1.78$ & \multicolumn{2}{c}{---} & \multicolumn{2}{c}{---} & \multicolumn{2}{c}{---} & 0.0141 & $+0.002$ \\
        Dragon LPIPS Hallstatt   & 0 & 0.9721 & $-0.005$ & $-1.15$ & \multicolumn{2}{c}{---} & \multicolumn{2}{c}{---} & \multicolumn{2}{c}{---} & 0.0125 & $+0.002$ \\
        Dragon LPIPS Skybox      & 0 & 0.8536 & $-0.108$ & $-2.44$ & \multicolumn{2}{c}{---} & \multicolumn{2}{c}{---} & \multicolumn{2}{c}{---} & 0.0353 & $+0.021$ \\
        Dragon ResNet50 Garden   & 0 & 0.8420 & $-0.151$ & $-2.44$ & 0.574 & $-0.108$  & 0.6927  & $-2.44$ & 0.5921  & $-2.44$ & 0.0855 & $+0.074$ \\
        Dragon ResNet50 Hallstatt & 0 & 0.8866 & $-0.103$ & $-2.44$ & 0.476 & $-0.073$  & 0.7621  & $-2.44$ & 0.6730  & $-2.44$ & 0.0660 & $+0.055$ \\
        Dragon ResNet50-SIN Skybox & 0 & 0.9974   & $-0.002$ & $-2.44$ & 0.377   & $-0.133$  & 0.9166   & $-2.44$ & 0.8782   & $-2.44$ & 0.0906 & $+0.076$ \\
        Dragon VGG Garden        & 0 & 0.9280  & $-0.034$ & $-2.24$ & \multicolumn{2}{c}{---} & \multicolumn{2}{c}{---} & \multicolumn{2}{c}{---} & 0.0152 & $+0.004$ \\
        Dragon VGG Hallstatt     & 0 & 0.9281  & $-0.016$ & $-1.38$ & \multicolumn{2}{c}{---} & \multicolumn{2}{c}{---} & \multicolumn{2}{c}{---} & 0.0122 & $+0.001$ \\
        Dragon VGG Skybox        & 0 & 0.7222  & $-0.186$ & $-2.44$ & \multicolumn{2}{c}{---} & \multicolumn{2}{c}{---} & \multicolumn{2}{c}{---} & 0.0255 & $+0.011$ \\
        Lion Statue CLIP Hallstatt & 0 & 0.6388   & $-0.214$ & $-2.44$ & 0.416    & $-0.104$  & 0.5975   & $-2.44$ & 0.4381   & $-2.44$ & 0.0825 & $+0.057$ \\
        Lion Statue CLIP Skybox  & 0 & 0.6008  & $-0.190$ & $-2.33$ & 0.359   & $-0.391$  & 0.5764   & $-2.44$ & 0.4234   & $-2.44$ & 0.1238 & $+0.102$ \\
        Lion Statue DINO Garden  & 0 & 0.6269  & $-0.290$ & $-2.44$ & 0.588    & $-0.242$  & 0.4893   & $-2.44$ & 0.4257   & $-2.44$ & 0.1049 & $+0.088$ \\
        Lion Statue DINO Hallstatt & 0 & 0.6077 & $-0.304$ & $-2.44$ & 0.476    & $-0.317$  & 0.4642   & $-2.44$ & 0.4040   & $-2.44$ & 0.0892 & $+0.063$ \\
        Lion Statue DINO Skybox  & 0 & 0.5843   & $-0.268$ & $-2.28$ & 0.492   & $-0.322$  & 0.4768   & $-2.44$ & 0.4138   & $-2.44$ & 0.1258 & $+0.103$ \\
        Lion Statue LPIPS Garden & 0 & 0.9571  & $-0.010$ & $-0.99$ & \multicolumn{2}{c}{---} & \multicolumn{2}{c}{---} & \multicolumn{2}{c}{---} & 0.0193 & $+0.002$ \\
        Lion Statue LPIPS Hallstatt & 0 & 0.9523  & $-0.011$ & $-0.95$ & \multicolumn{2}{c}{---} & \multicolumn{2}{c}{---} & \multicolumn{2}{c}{---} & \cellcolor{green!20}0.0244 & $-0.002$ \\
        Lion Statue LPIPS Skybox & 0 & 0.9230  & $-0.014$ & $1.96$ & \multicolumn{2}{c}{---} & \multicolumn{2}{c}{---} & \multicolumn{2}{c}{---} & 0.0237 & $+0.001$ \\
        Lion Statue ResNet50 Garden & 0 & 0.8841    & $-0.092$ & $-2.21$ & 0.599  & $-0.251$  & 0.6188   & $-2.44$ & 0.5190    & $-2.44$ & 0.1209 & $+0.104$ \\
        Lion Statue ResNet50 Hallstatt & 0 & 0.8907    & $-0.085$ & $-2.29$ & 0.521    & $-0.232$  & 0.6174    & $-2.44$ & 0.5071    & $-2.44$ & 0.0952 & $+0.069$ \\
        Lion Statue ResNet50 Skybox & 0 & 0.8786   & $-0.081$ & $-2.09$ & 0.222  & $-0.578$  & 0.6553   & $-2.44$ & 0.5572   & $-2.44$ & 0.1243 & $+0.102$ \\
        Lion Statue ResNet50-SIN Garden & 0 & 0.9976  & $-0.001$ & $-2.44$ & 0.528 & $-0.063$  & 0.9214    & $-2.44$ & 0.8843    & $-2.44$ & 0.1215 & $+0.105$ \\
        Lion Statue ResNet50-SIN Hallstatt & 0 & 0.9978   & $-0.001$ & $-2.44$ & 0.503   & $-0.015$  & 0.9267   & $-2.44$ & 0.8922   & $-2.44$ & 0.0960 & $+0.070$ \\
        Lion Statue ResNet50-SIN Skybox & 0 & 0.9975  & $-0.001$ & $-2.44$ & 0.434    & $-0.244$  & 0.9200   & $-1.38$ & 0.8821   & $-2.10$ & 0.1224 & $+0.100$ \\
        Lion Statue VGG Garden   & 0 & 0.9078  & $-0.014$ & $-0.76$ & \multicolumn{2}{c}{---} & \multicolumn{2}{c}{---} & \multicolumn{2}{c}{---} & 0.0191 & $+0.002$ \\
        Lion Statue VGG Skybox   & 0 & 0.7432  & $-0.106$ & $-0.24$ & \multicolumn{2}{c}{---} & \multicolumn{2}{c}{---} & \multicolumn{2}{c}{---} & 0.0238 & $+0.002$ \\
        Suzanne ResNet50 Skybox  & 0 & 0.8852  & $-0.079$ & $-1.45$ & 0.743    & $-0.105$  & 0.6599  & $-1.78$ & 0.5602  & $-1.86$ & 0.0480 & $+0.033$ \\
        \bottomrule
    \end{tabular}
\end{table*}

\paragraph{Environment and Shape Effects}
All three lighting conditions in our shape experiments --- \textit{Garden}, \textit{Hallstatt}, and \textit{Skybox} --- are high-dynamic-range (HDR) environment maps; they differ not in whether they are HDR but in the spatial frequency and contrast of their illumination, with Skybox carrying the most high-frequency specular content. We use ``environment map'' for this factor throughout. Across all shape experiments, the environment map has a systematic and geometry-dependent effect on recoverability. Skybox illumination is the most challenging condition, yielding metamers (here counting experiments that meet at least one criterion) in only 8 of 24 experiments (33\%), compared to 13 of 24 under Garden (54\%) and 12 of 24 under Hallstatt (50\%). This difficulty is partially explained by the lower baseline cosine similarities under Skybox (mean $0.921$ vs.\ $0.966$--$0.965$ for Garden and Hallstatt), indicating that even the ground-truth image-space reconstruction produces a weaker latent signal under high-frequency environment maps --- the optimization landscape itself is harder. The effect is most pronounced for geometrically complex objects: The Lion Statue produces no metamers under Skybox for any representation, and Dragon yields only one (ResNet50 Skybox). By contrast, Suzanne is largely invariant to the choice of environment map (5--6 metamers across all three), confirming that geometric simplicity dominates over lighting difficulty. A few representations even exhibit ``inverted'' HDR sensitivity: VGG succeeds on Dog Skybox ($\Delta_{\text{sim}} = +0.093$) while failing in Garden and Hallstatt, and ResNet50 produces a metamer on Dragon Skybox while falling short on Garden and Hallstatt. This suggests that high-frequency specular content introduced by the Skybox can, in specific cases, provide richer gradient signal that outweighs the increased optimization complexity.

\begin{figure}
    \centering
    \includegraphics[width=\linewidth]{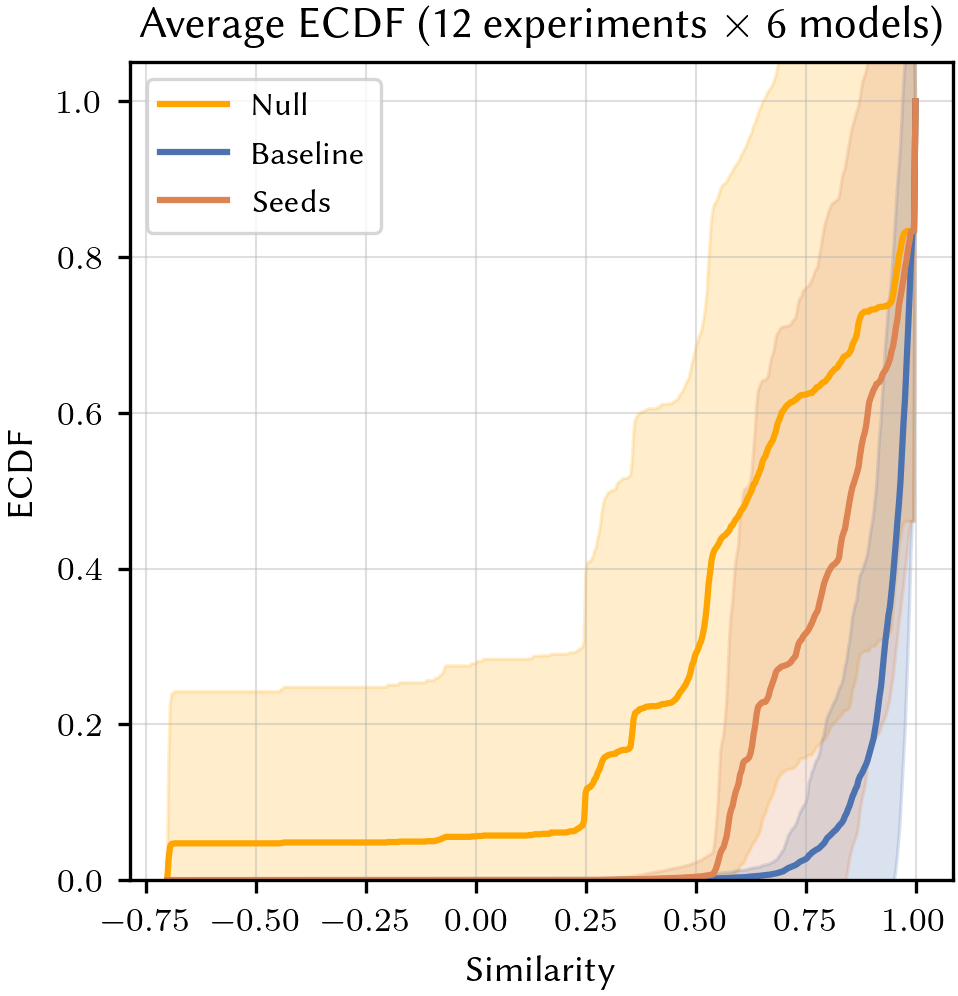}
    \caption{Mean empirical cumulative distribution functions (ECDFs) of cosine similarity across all shape experiments (three environment maps, four geometries, six models). The three curves are the control distributions that calibrate our metamer criterion, spanning from chance to the attainable ceiling. (a) The null distribution (orange): similarity to the target obtained from 400 random object meshes unrelated to the target --- a chance floor for what counts as ``similar.'' (b) The seeds distribution (teal): similarity across independent reconstruction runs of the same scene with different random initializations --- a measure of run-to-run stability. (c) The baseline distribution (purple): the similarity achieved by the image-space ground-truth reconstruction used for hyperparameter selection (the baseline does not use the network in its loss) --- the attainable ceiling. The ordering null $\ll$ seeds $\lesssim$ baseline is what justifies requiring reconstruction similarity to reach the baseline before declaring a metamer. Higher similarity denotes closer latent alignment; shaded regions are one standard deviation.}
    \label{fig:ecdf}
\end{figure}

\paragraph{Empirical Cumulative Distributions as Criterion Calibration}
Figure~\ref{fig:ecdf} shows the average empirical cumulative distribution functions (ECDFs) of cosine similarity across all shape experiments under three conditions. The null distribution, computed from 400 randomly perturbed object meshes, occupies the lowest region of the similarity axis but exhibits a notably wide spread ($\sigma \approx 0.21$--$0.49$ across shapes), with a long right tail reaching near-perfect similarity at its 95th percentile ($\geq 0.997$). This tail reflects the fact that some random objects happen to share incidental structural features with the target in a given representation's latent space, which motivates calibrating the metamer criterion against a principled reference rather than against zero or against chance alone. The seeds distribution, obtained from independent re-runs of each experiment with different random initializations, is tightly concentrated well above the null (mean $0.87$--$0.93$, $\sigma \approx 0.08$--$0.14$), demonstrating that the optimization converges to a stable region of representation space regardless of initialization. The baseline distribution lies at the upper end of the similarity spectrum (mean $0.92$--$0.96$), reflecting the latent alignment achievable when a reconstruction is driven directly towards the ground-truth scene in image space --- recall that the baseline does not use the network in its loss (Section~\ref{sec:baseline}); the network is used only to read out the resulting latent similarity. The clear stacking $\text{null} \ll \text{seeds} \lesssim \text{baseline}$ validates the metamer criterion: by requiring reconstruction similarity to reach or exceed the baseline, we set a threshold that is reliably above both random similarity and seed-induced variation, while remaining attainable for representations that encode scene-relevant information faithfully. The tight seeds distribution further confirms that the pass/fail outcome is a property of the representation --- scene pairing rather than an artefact of optimization stochasticity.

\subsection{Additional analyses}\label{sec:latent-structure}
The per-criterion correlation and RSA values are reported per object in Sections~\ref{sec:results-material} and~\ref{sec:results-shape}. Here we add three analyses that those per-experiment tables do not make obvious: a direct comparison of standard and shape-biased ResNet, generalization to a held-out view, and a Bayes-factor calibration of the binary criterion. (We keep the high-level interpretation for the Discussion.)

\begin{figure*}
    \centering
    \includegraphics[width=\linewidth]{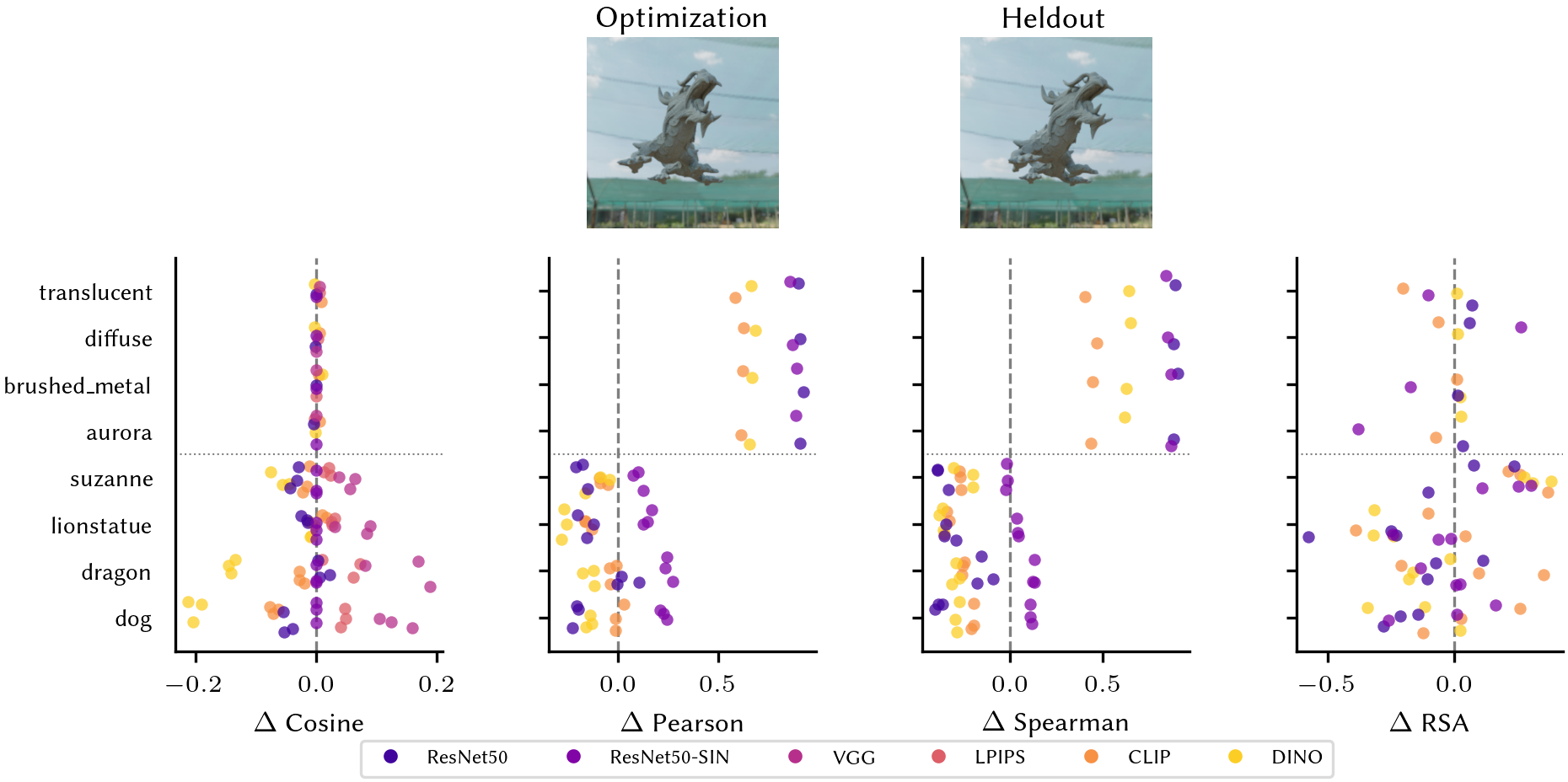}
    \caption{Held-out-view generalization. Each dot shows the difference between the peak reconstruction similarity and the corresponding baseline model similarity on a held-out view (reconstruction $-$ baseline), for one experiment. Dots to the right of the dashed zero line indicate that the reconstruction matches or exceeds the baseline model's own response to the target scene on an unseen viewpoint. Color encodes the perceptual network used as the optimization objective. The dotted horizontal rule separates material-optimization experiments (top) from shape-optimization experiments (bottom).}
    \label{fig:heldout}
\end{figure*}

\paragraph{Standard versus shape-biased ResNet} ResNet-SIN was included to test whether its shape-biased training (on Stylized ImageNet) changes what MRD recovers --- the hypothesis being that a more shape-biased network should be easier to drive to a shape metamer. It does change the representation, but not in that direction. Across both materials and shapes, ResNet-SIN yields consistently weaker RSA than standard ResNet (e.g.\ material RSA $0.42$--$0.86$ vs.\ $\geq 0.997$), even though its cosine similarity on shapes is near-perfect ($\approx 0.997$) while its Pearson and Spearman plateau lower ($\approx 0.92$ and $\approx 0.88$). The shape bias thus aligns global feature directions without pinning down the internal, inter-view structure: SIN training measurably altered the representation, but not in a way that improves shape metamerism here.

\paragraph{Held-out-view generalization} To assess whether the optimized reconstructions generalize beyond the training views, we evaluated each metric on a single held-out view withheld from the objective during optimization (Figure~\ref{fig:heldout}). For material scenes, held-out cosine similarity was near parity with the rendered baseline across all networks (mean $\Delta = +0.001$, range $[-0.004, +0.009]$), while held-out Pearson and Spearman correlations were consistently and substantially positive (Pearson $\Delta \in [+0.58, +0.93]$; Spearman $\Delta \in [+0.41, +0.90]$), indicating that the optimized material properties transfer well to unseen illumination conditions. For shape scenes the picture is more network-dependent: VGG and LPIPS reconstructions retained a positive cosine advantage on the held-out view (mean $\Delta \approx +0.11$ and $+0.04$, respectively), whereas DINO-optimized reconstructions showed a consistent deficit ($\Delta \in [-0.13, -0.21]$), suggesting that DINO features are sensitive to view-dependent appearance that the geometry optimization cannot fully reproduce. ResNet50 on SIN occupied an intermediate position, with cosine differences near zero but consistently positive Pearson and Spearman held-out differences ($\Delta \approx 0.10$--$0.27$). RSA differences on the held-out view were mixed across both groups ($50\%$ positive overall), indicating no systematic advantage or deficit in representational geometry at the novel viewpoint.

\paragraph{Bayes factors as evidence calibration}
The rank-based Bayes factor BF$_{+0}$ largely corroborates the binary evaluation criteria, agreeing in 90 of 96 experiments (94\%), but the six disagreeing cases are instructive. In three material experiments --- Aurora ResNet50, Diffuse ResNet50, and Diffuse ResNet50-SIN --- the binary criterion is satisfied ($\Delta_{\text{sim}} > 0$) yet the Bayes factor favors $H_0$ ($\log_{10}\text{BF} \approx -0.06$ to $-0.14$). This arises because the per-view reconstruction similarities cluster so tightly around the baseline that the rank-based test perceives no reliable directional shift; the reconstruction is representationally equivalent to the baseline without exhibiting a consistent improvement across views. In the opposite direction, Dog VGG Garden and Lion Statue LPIPS Skybox miss the binary criterion by $\Delta_{\text{sim}} \approx -0.002$ and $-0.014$ respectively, yet carry moderate-to-strong Bayes factor evidence for $H_+$ ($\log_{10}\text{BF} = 3.62$ and $1.96$). Here, the per-view similarity distribution is consistently shifted toward the baseline even though the single peak epoch falls marginally short. More broadly, shape non-metamers cluster at the lower bound of the BF scale ($\log_{10}\text{BF} \approx -2.44$ in 60 of 72 cases), indicating unambiguous and consistent failure rather than near-misses. The strongest positive evidence concentrates in LPIPS and VGG runs on simpler geometries ($\log_{10}\text{BF}$ up to 13.2 for Dog VGG Skybox), where the reconstruction consistently and reproducibly exceeds the baseline across views. Thus near-threshold binary outcomes should be taken with skepticism: a marginal positive $\Delta_{\text{sim}}$ without supporting BF evidence may reflect single-epoch noise rather than genuine representational equivalence.

\section{Discussion}\label{sec:discussion} 

Our central result is a proof of concept: making physically based differentiable rendering converge against a deep-network objective is a hard technical problem, and within that constraint MRD reliably finds material metamers but only occasionally finds shape metamers. The relative success for material is a success of MRD, not evidence that the networks recover materials accurately: the metamers MRD finds are often not the ground-truth material, which tells us the networks' material equivalence classes are broad. 
The relative difficulty for shape is genuine --- more often than not a shape reconstruction meets none of our metamer criteria --- but there are clear success cases (perceptual losses, and simple geometries such as Suzanne) that demonstrate the method can work. The rest of this section interprets why this asymmetry arises, discusses the relationship between RSA and point-wise similarities, situates MRD relative to rasterization-based probing, and argues what a richer ``concept-level'' reconstruction would require.

\paragraph{Why Material Reconstruction Outperforms Shape Reconstruction}
The gap between material and shape is large and consistent across every metric we report (Section~\ref{sec:results}). We attribute this asymmetry to four interacting causes. (1) Material variation occupies a substantially lower-dimensional and more structured subspace than shape: changes in roughness, metallicity, or anisotropy produce smooth, predictable changes in shading that map roughly linearly into a network's latent space, whereas small geometric perturbations induce large, non-linear changes in silhouette, occlusion, and local shading, so the latent response is high-dimensional and hard to match coordinate-wise. (2) Material edits affect the image globally and therefore produce spatially dense, coherent gradients, whereas shape edits produce sparse or discontinuous gradients concentrated at silhouette boundaries, which slows convergence. (3) Modern vision networks encode texture and shading statistics more explicitly than geometry~\cite{geirhosIMAGENETTRAINEDCNNSARE2019}, so their latents are intrinsically more sensitive to material than to shape. 
ResNet-SIN makes this visible in an extreme form, where its shape bias saturates cosine similarity for almost any shape while leaving the structural metrics underspecified. (4) The material optimization landscape is smoother, with lower intra-class variance in the baseline embeddings. The convex, low-detail Suzanne is the clearest support for this account: it is the one shape object whose relational structure exceeds the baseline in the majority of conditions, suggesting that once geometric complexity is reduced, shape behaves more like material. Importantly, the strong material numbers should still be read as evidence about MRD and about the breadth of the networks' equivalence classes, not as evidence that the networks recover the true material parameters.

\paragraph{Relationship between RSA and Point-wise Similarities} 
Our shape experiments identify interesting dissociations between performance measured by point-wise similarities (e.g. measured with cosine similarity) and RSA.
This highlights an interesting property of these metrics, that per-view cosine similarity and RSA measure structurally orthogonal properties of the recovered representations. 
Per-view cosine similarity asks: for each individual view, how close is the recovered latent to the ground-truth latent? 
RSA asks: does the recovered set of latents preserve the pairwise relational geometry of the ground-truth set? 
High accuracy on the first does not tightly constrain the second. To see why, consider a systematic error mode in which the optimizer recovers per-view latents that are each individually close to their ground-truth counterparts, but where the error vectors across views are correlated --- for example, all recovered latents are slightly contracted toward the mean representation of the object. Each individual cosine similarity remains very high (errors are small in magnitude), but inter-view pairwise distances are uniformly compressed. Because RSA is a rank-order correlation over the upper triangle of the RDM, this compression can reorder the fine-grained distance structure substantially, even when the absolute magnitude of the distortion is small.
This sensitivity is especially pronounced for ResNet50-SIN precisely because the network is shape-biased and highly view-invariant: ground-truth latents for different views of the same object cluster tightly together, giving a ground-truth RDM with very low variance. When the dynamic range of the true RDM is compressed into a narrow interval, even small absolute errors in recovered inter-view distances are sufficient to scramble their rank order. In our updated results, ResNet50-SIN achieves a mean pointwise similarity of 0.998 across all 16 conditions, yet yields a mean RSA of 0.540 --- lower than DINO (RSA 0.616 at mean similarity 0.728) and ResNet50 (RSA 0.621 at mean similarity 0.900), both of which have richer inter-view geometry that is coarser but more rank-stable. This pattern --- lower-similarity networks yielding higher RSA — is precisely what one would predict from this sensitivity argument.

\paragraph{Physically Based Light Transport vs. Rasterization} Recently, Elumalai et~al. \cite{elumalai2025beyond} optimized a mesh to obtain maximally exciting inputs for a Robust ResNet's activation using a differentiable rasterizer. In their work, they successfully use differentiable rendering to optimize against a network activation objective. 
However, rasterization lacks physicality, as it is bound to approximations of global illumination and/or screen-space techniques. In contrast, MRD reconstructs latent representations using a differentiable path tracer, thereby accounting for physically based light transport and simulating light propagation in a 3D scene.

\paragraph{Concepts in Higher Layers}
Finally, we note that most of the qualitative visualizations we presented here imply that a successful reconstruction is one that appears subjectively similar to the target image (and indeed, that is how we quantify this result in Figure \ref{fig:two-outcomes}).
However, depending on the model and the representational layer(s) used in the reconstruction, this is not necessarily the only possible ``success''.
Consider that Feather and colleagues \cite{featherModelMetamersReveal2023} demonstrated that matching the activations of early layers of a convolutional neural network can produce similar images to the target, whereas later layers exhibit images that appear essentially meaningless to humans.
We speculate that MRD, due to its physical basis and constraints, may allow researchers to gain insight into the nature of the semantic-level representation of a given network: what features does the model learn to become invariant to in representing a higher-level concept?
Consider that asking a human to visualize a ``dragon" would likely elicit rich shape-based descriptions. A person would not think of exactly the target dragon, but of other forms that other humans would also consider to be a dragon.
We assert that, were it possible to run MRD on a human, the result would be a series of dragon-like meshes that other observers would likewise classify as ``dragons". Not all of the same dragon, and not all in the same pose, but dragons nonetheless.
If a network existed that had a human-like visual semantic representation of dragons (including shape-based descriptors), then we might expect MRD to reconstruct a family of dragon-like shapes.
One interpretation of our present results is that such a network currently does not exist.

\subsection{Future work}
We intentionally chose simple scenes to demonstrate the utility of MRD as a proof of concept. Future work could study other factors (camera position, lighting) and also combinations of scene parameters together. A small but concrete improvement would be to place the object on a ground plane rather than floating it, so that contact shadows and indirect bounces provide additional, physically grounded cues for the optimization.
Furthermore, more computationally expensive models can be probed. PBDR is an active field in computer graphics, and advances in PBDR will help to improve future iterations of MRD. In particular, while our shape results more often than not do not meet any criterion for metamerism, integrating recent advances such as \textit{Many Worlds Rendering} \cite{zhangManyWorldsInverseRendering2024} --- which better handles the discontinuous, high-variance gradients that dominate at silhouette boundaries --- may improve the robustness of MRD for complex mesh reconstruction.

Our work has additional technical limitations. First, probing larger NNs is non-trivial; we require sharding the model across multiple GPUs. When Mitsuba is involved, inter-process communication is required to send gradients between devices so we can run the backward pass across GPUs. Second, rendered images may represent an out-of-distribution problem for the models we tested; fine-tuning on renderings may improve results.
Finally, because we test our method only on rendered images, one reason the models perform so poorly in shape reconstruction may be that they are not, in general, trained on our scenes. Our scenes view objects from unusual views below the object, while in ImageNet, objects are primarily viewed from above or the side.

\section{Conclusion}
We present \textbf{MRD} (metamers rendered differentiably), a novel approach to reasoning about image-computable models in physically grounded 3D space. By optimizing physical scene parameters against a frozen network's latent representation, MRD can reconstruct model metamers, letting us ask whether a model is invariant or responsive to a given physical scene parameter (such as material or shape). As a proof of concept, MRD reliably finds material metamers --- though these are frequently not the ground-truth material, indicating broad equivalence classes --- while shape reconstructions more often than not meet none of our metamer criteria, with clear success cases under perceptual losses and simple geometries. Getting physically based differentiable rendering to converge against a deep-network objective is itself a substantial technical challenge, and these results establish both that the approach is viable and where it currently breaks down. As the first work in this direction, there are significant limitations and clear opportunities for improvement, particularly through advances in differentiable rendering that stabilize the gradients governing complex geometry.

\section*{Author contributions}
The original idea to apply PBDR as an evaluation method to understand learned visual representations was developed jointly by BB and TSAW.
BB performed all implementation and analysis.
BB and TSAW jointly wrote and edited the paper.

\section*{Acknowledgments}
We thank Joshua Martin for feedback on the manuscript. We further thank the following artists and authors for providing the assets for non-commercial purposes: \textit{Delatronic (Material Dragon), Stanford's Dragon model (Shape Dragon), Rigsters' Lion crushing a serpent (Lion Statue), and Blender's Suzanne. For the Hallstatt envmap, we thank Bernhard Vogl. The other three environment maps were available on HDRI Haven. The Aurora and Brushed Metal materials were measured and provided by Dupuy and Jakob \cite{Dupuy2018Adaptive}.}

\section*{Disclosure of Machine Learning Tools Use}
In the preparation of this manuscript, we use large language models (LLM) for orthographic purposes, such as the correction of spelling and grammar and minor rephrasing.

\section*{Funding}
Funded by the European Union (ERC, SEGMENT, 101086774). Views and opinions expressed are, however, those of the author(s) only and do not necessarily reflect those of the European Union or the European Research Council. Neither the European Union nor the granting authority can be held responsible for them.
This work was additionally supported by the Deutsche Forschungsgemeinschaft (German Research Foundation, DFG) under Germany’s Excellence Strategy (EXC 3066/1 “The Adaptive Mind”, Project No. 533717223).

\bibliographystyle{ACM-Reference-Format}


\end{document}